\documentclass{article} %
\usepackage{colm2024_conference}
\usepackage{booktabs}
\usepackage{graphicx}
\usepackage{enumitem}
\usepackage{wrapfig}
\usepackage{algorithm}
\usepackage{algpseudocode}
\usepackage{wrapfig}
\usepackage{float}
\usepackage{microtype}
\usepackage{amsmath}
\usepackage{amssymb}
\usepackage{colortbl}
\usepackage[T1]{fontenc}
\usepackage[utf8]{inputenc}
\usepackage{textcomp}
\definecolor{lightgray}{rgb}{0.9,0.9,0.9}
\usepackage{caption}
\usepackage{subcaption}
\usepackage{xcolor}
\usepackage{setspace}
\usepackage{url}
\usepackage{multirow}
\usepackage{colortbl}
\usepackage{tabularx}
\usepackage{blindtext}
\usepackage{pgfplots}
\pgfplotsset{compat=1.18} 
\usepackage{tikz}
\usetikzlibrary{
  arrows.meta,
  calc,
  er,
  positioning,
  bayesnet
}
\usepackage{makecell}
\usepackage[safe]{tipa}
\usepackage{siunitx}
\usepackage{nicefrac}
\usepackage{tocloft}
\usepackage{listings}
\usepackage[raster,skins]{tcolorbox} %
\usepackage{xltabular}
\usepackage{adjustbox}
\usepackage{xurl}
\usepackage{graphicx}
\usepackage{mwe}

\usepackage{amsmath,amsfonts,bm}

\def\eqref#1{equation~\ref{#1}}

\def\1{\bm{1}}

\def\vs{{\bm{s}}}

\DeclareMathAlphabet{\mathsfit}{\encodingdefault}{\sfdefault}{m}{sl}
\SetMathAlphabet{\mathsfit}{bold}{\encodingdefault}{\sfdefault}{bx}{n}

\usepackage{tikz}
\usetikzlibrary{arrows.meta,calc,positioning}

\makeatletter
\DeclareRobustCommand\onedot{\futurelet\@let@token\@onedot}
\def\@onedot{\ifx\@let@token.\else.\null\fi\xspace}

\def\eg{\emph{e.g}\onedot} 
\def\ie{\emph{i.e}\onedot} 
 
\def\etc{\emph{etc}\onedot} \def\vs{\emph{vs}\onedot}

\lstdefinestyle{captionjson}{
    basicstyle=\ttfamily\scriptsize,
    backgroundcolor=\color{gray!5},
    frame=single,
    rulecolor=\color{gray!35},
    breaklines=true,
    breakatwhitespace=false,
    columns=fullflexible,
    keepspaces=true,
    showstringspaces=false,
    tabsize=2,
    xleftmargin=6pt,
    xrightmargin=6pt,
    framexleftmargin=6pt,
    framexrightmargin=6pt,
    aboveskip=0pt,
    belowskip=0pt
}

\makeatother

\newcommand{\shortname}{\xspace Hy-Nomas\xspace}

\title{\shortname: Towards Free-form Interactive World Generation}
\author{
\bf Tencent Hy Visual Generation Team
}

\begin{document}
\maketitle
\begin{abstract}
Interactive world generation requires models to follow successive commands while preserving relevant history and action consequences. We present\textbf{\shortname}, a video world model with an 8B diffusion backbone that supports free-form text instructions and continuous camera control across multiple interaction turns. It generates variable-length segments conditioned on current inputs and cached history. Our central design aligns structured video captions with action-aware causal attention: descriptive context remains accessible across segments, while action instructions are restricted to their intended temporal windows. A vision-language Prompt Enhancer (PE) translates user requests and multimodal interaction histories into structured conditions for the next segment. A multi-stage curriculum progressively introduces causal generation, text interaction, camera control, and training on model-generated histories, followed by diffusion distillation. In human evaluation, \shortname shows superior performance over existing world models. We release model weights, code and benchmarks, together with a lightweight adaptation workflow, to support reproducible research on interactive world generation.
\end{abstract}

\begingroup
\setlength{\intextsep}{5pt}
\begin{figure}[!ht]
\centering
\includegraphics[width=\linewidth]{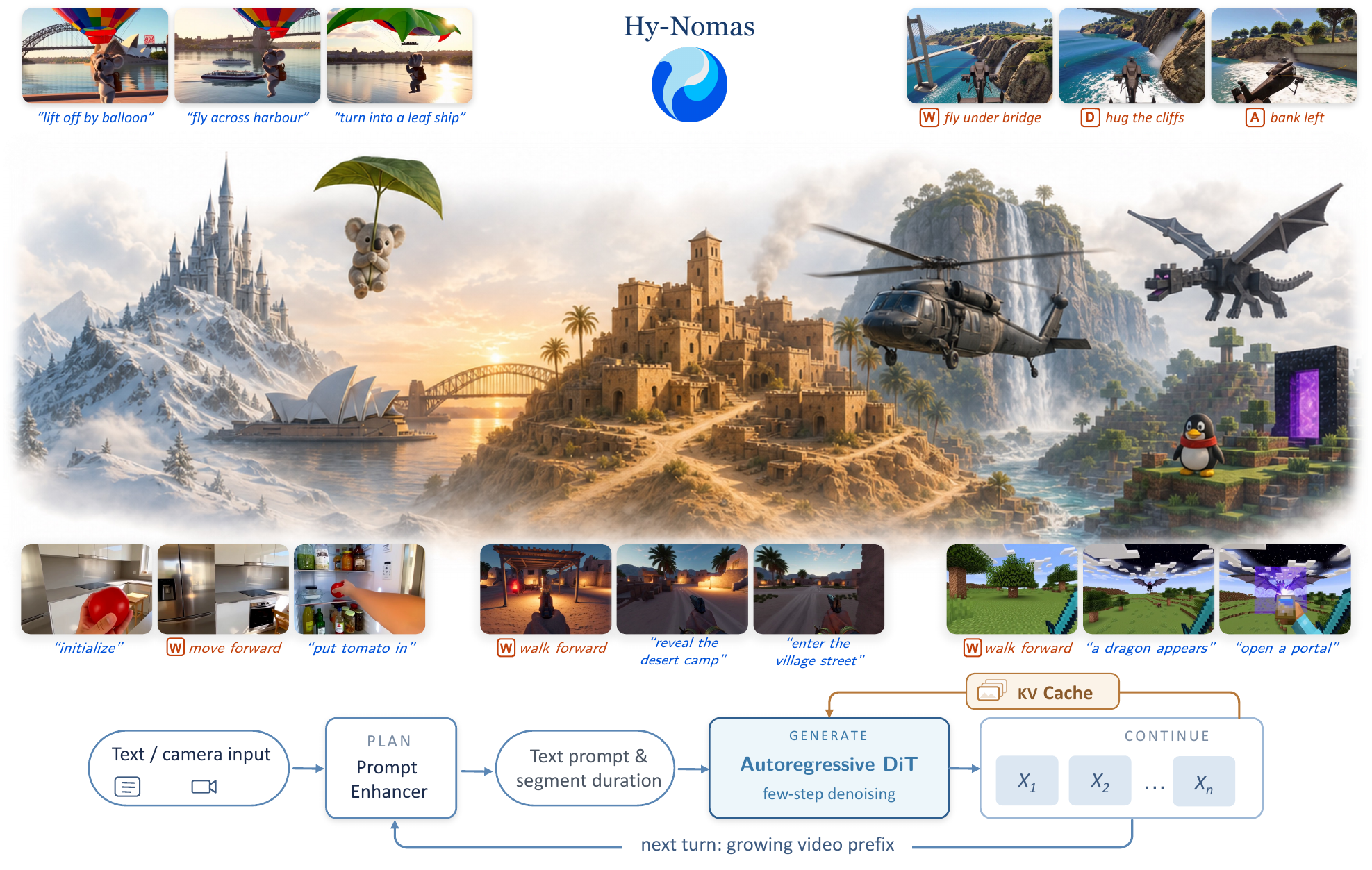}
\captionsetup{font=small,skip=4pt}
\caption{\textbf{Interactive world generation with \shortname{}.} Free-form text instructions and continuous camera control can be interleaved across successive interaction turns. At each turn, a vision-language Prompt Enhancer converts the user request and observed world history into a structured text prompt and a duration for the next video segment. Conditioned on these inputs and cached context, \shortname{} autoregressively generates the next segment, allowing the world to unfold through continued interaction.}
\label{fig:teaser}
\end{figure}
\endgroup

\clearpage
\setcounter{tocdepth}{2}
\tableofcontents
\clearpage

\section{Introduction} 
We define a video world model as an interactive virtual environment that renders visual observations of future states in response to user commands. Users observe the current world, decide how to act, and the model generates new visual observations that reflect both their commands and the preceding interaction history. This closed interaction loop extends the objective of conventional video generation, where a fixed prompt typically specifies a short video clip in advance. An interactive world must accommodate input commands that arrive during generation, model action-dependent dynamics, and preserve the consequences of earlier actions.
Recent systems, including Genie 3~\citep{deepmind2025genie3}, HappyOyster~\citep{hyintro_happyoyster_2026} and LingBot-World v2~\citep{gao2026infinite}, have made substantial progress in navigation, semantic intervention, and long-horizon generation~\citep{bolton2025sima,bar2025navigation,yang2023unisim}. Building on these advances, we introduce \shortname, an 8B-parameter world model focused on execution of free-form text instructions and their integration with camera navigation across multi-turn interactions. \shortname enables users to direct both what happens in a world and how they explore it through one generative backbone.

Building such a model requires more than simply replacing full attention with a causal mask. A pretrained bidirectional diffusion transformer (DiT)~\citep{peebles2023scalable} must be adapted to generate successive segments from available history while following the current user command, preventing information leakage that breaks causality, and supporting efficient reuse of cached context~\citep{hyintro_causvid_2025,hyintro_causalforcing_2026,hyintro_minwm_2026}. Because each generated segment becomes evidence for the next, small errors can compound into appearance drift, implausible motion, and scene breakdown; training must therefore prepare the model for its own imperfect predictions~\citep{huang2025selfforcing,hyintro_longlive_2025,oasis2024}. Multi-turn interaction adds a second difficulty: persistent scene information must remain accessible, while an action instruction should take effect only during its intended time interval~\citep{hyintro_h3world_2026,mao2025yume15,hyintro_worldplay2_2026,nvidia2025cosmos}. The model must also coordinate free-form semantic commands with geometric camera controls, and retain identities, layouts, and state changes when previously observed content returns into view~\citep{he2024cameractrl,xiao2025worldmem,yu2025contextmemory}. Finally, temporally causal information flow alone does not establish physical causality. Learning plausible action-to-state transitions requires training data that connects actions with their observable consequences and evaluation that tests these behaviors across successive interactions~\citep{ying2026wbench,xu2026worldroambench,bai2026happyworldbench}.

\shortname can generate visual worlds with different styles and viewpoints, supporting minute-scale rollouts with successive instructions for subject actions, object interactions, scene changes and camera motion, as shown in Figure~\ref{fig:teaser}. Free-form text prompts and camera control commands can be interleaved within the same session, with historical visual and semantic context carried forward through the autoregressive key-value (KV) cache. A vision-language model (VLM), the Prompt Enhancer (PE)~\citep{qwen38} converts the user's request and current observations into a structured prompt with an appropriate duration for the next segment. Our evaluation demonstrates that \shortname is superior to existing world models across a range of interactive tasks.

We build \shortname by co-designing our data strategy, video captioning, model architecture, and training strategy towards free-form interactive world generation. Our data pipeline sources videos with physical interaction, human and object motion, camera movement, and long-range dependencies; video captioning records persistent subject identities, scene changes, and timestamped actions, complemented by estimated camera trajectories~\citep{huang2025vipe}. The model architecture design couples these annotations with action-aware causal attention. Descriptive text remains available across subsequent segments, while action-specific text is restricted to its originating segment and annotated time window. This separation is designed to retain descriptive context and the visual consequences of an action without repeatedly reissuing the action itself. Mixed text-only and camera-conditioned training supports transitions between two interaction modes. Starting from HunyuanVideo 1.5~\citep{wu2025hunyuanvideo}, we progressively introduce segment causality, localized action conditioning, and camera control, followed by high-quality annealing and segment forcing into the model training process. To make this training practical, we combine a high-compression video autoencoder~\citep{chen2025dc}, FLOPs-balanced packing of variable-length videos~\citep{krell2022packing}, attention kernels that execute heterogeneous masks~\citep{magiattention2025}, distributed sequence and model-state sharding~\citep{jacobs2023deepspeed,pytorch_fsdp2}, and stage-dependent memory management, supporting packed sequences of up to $360K$ tokens. Bounded-cache inference and a LoRA workflow that transfers adapters from the base model to its distilled counterpart further support deployment and customization for downstream applications.

Our main contributions are:
\begin{itemize}
[leftmargin=15pt]
    \item \textbf{An interactive 8B world model.} We develop \shortname for multi-turn free-form interaction and continuous camera navigation within one autoregressive DiT.

    \item \textbf{Aligned supervision and action conditioning.} We combine structured chronological captions, persistent subject descriptions, timestamped actions,
and camera trajectories with action-aware causal attention that distinguishes enduring context from temporally localized
instructions.

    \item \textbf{A scalable training recipe and open-weight model.} We integrate a progressive strategy to gradually extend the model's capability, together with bounded-cache inference and lightweight downstream adaptation. We release model weights and code to support further research.

    \item \textbf{Evaluation of world-model capabilities.} We assess multi-turn text interaction, camera navigation, interleaved control, and spatial memory, demonstrating the advantages of \shortname while identifying the remaining challenges in world modeling. We release all the benchmarks to facilitate reproducible research on world-model evaluation.
\end{itemize}

\section{Data}
\begin{figure*}[t]
  \centering
  \resizebox{\linewidth}{!}{%
\begingroup%
\definecolor{dpInk}{HTML}{263746}%
\definecolor{dpMuted}{HTML}{667583}%
\definecolor{dpBlue}{HTML}{347AA0}%
\definecolor{dpAmber}{HTML}{A57829}%
\definecolor{dpPurple}{HTML}{7955A0}%
\definecolor{dpRose}{HTML}{A45F67}%
\definecolor{dpTeal}{HTML}{238476}%
\definecolor{dpSampling}{HTML}{52648C}%
\definecolor{dpLine}{HTML}{D6DFE5}%
\newcommand{\dpText}[5]{%
  \node[anchor=north west,inner sep=0,outer sep=0,text=#4,%
    align=left,dpFont={#3}{18}{12}]%
    at (#1,#2) {#5};}%
\newcommand{\dpCard}[7]{%
  \filldraw[fill=white,draw=#5!48,line width=.8bp,rounded corners=4bp]%
    (#1,#2) rectangle (#3,#4);%
  \begin{scope}%
    \clip[rounded corners=4bp] (#1,#2) rectangle (#3,#4);%
    \fill[#5!9] (#1,#2) rectangle (#3,{#2+34});%
  \end{scope}%
  \node[anchor=west,inner sep=0,text=#5,%
    dpFont={15.5}{18}{12}]%
    at ({#1+10},{#2+17}) {\strut\textbf{#6\enspace #7}};}%
\begin{tikzpicture}[x=1bp,y=-1bp,
  dpFont/.style n args={3}{
    font={\pgfmathsetmacro{\dpLeading}{#2*#3/#1}%
      \sffamily\fontsize{#3}{\dpLeading}\selectfont},
    scale={#1/#3}},
  flow/.style={draw=dpMuted,line width=1bp,-{Stealth[length=4bp,width=3bp]},
    rounded corners=3bp}]

\fill[dpInk!6,rounded corners=3bp] (0,0) rectangle (836,32);
\dpText{11}{7}{15}{dpInk}{\mbox{\textbf{INPUT VIDEOS}}}
\dpText{146}{7}{15}{dpInk}{\mbox{Real and virtual environments}}
\draw[flow] (40,32)--(40,48);

\dpCard{0}{48}{200}{250}{dpBlue}{01}{Preprocessing}
\dpCard{214}{48}{390}{250}{dpAmber}{02}{Data distribution}
\dpCard{404}{48}{604}{250}{dpPurple}{03}{Video tagging}
\dpCard{618}{48}{836}{250}{dpInk}{04}{Training stages}
\foreach \a/\b in {200/214,390/404,604/618}{
  \draw[flow,shorten <=2bp,shorten >=2bp] (\a,65)--(\b,65);}

\node[anchor=west,inner sep=0,outer sep=0,text=dpInk,align=left,
  dpFont={15}{19}{12}]
  at (16,115) {\mbox{Split \& transcode}};
\node[anchor=west,inner sep=0,outer sep=0,text=dpInk,align=left,
  dpFont={15}{19}{12}]
  at (16,152) {\mbox{Video property filtering}};
\node[anchor=west,inner sep=0,outer sep=0,text=dpMuted,align=left,
  dpFont={14}{17}{10}]
  at (16,184) {\mbox{Resolution, aspect ratio,}\\\mbox{duration, etc.}};
\node[anchor=west,inner sep=0,outer sep=0,text=dpInk,align=left,
  dpFont={15}{19}{12}]
  at (16,230) {\mbox{Visual quality filtering}};

\node[anchor=west,inner sep=0,outer sep=0,text=dpInk,align=left,
  dpFont={15}{19}{12}]
  at (226,115) {\mbox{Hierarchical clustering}};
\draw[draw=dpAmber!45,line width=.6bp,rounded corners=2bp]
  (228,129)--(228,152)--(234,152);
\node[anchor=west,inner sep=0,outer sep=0,text=dpInk,align=left,
  dpFont={15}{19}{12}]
  at (238,152) {\mbox{Within-cluster dedup}};
\node[anchor=west,inner sep=0,outer sep=0,text=dpInk,align=left,
  dpFont={15}{19}{12}]
  at (226,230) {\mbox{Taxonomy labeling}};

\filldraw[fill=white,draw=dpTeal!45,line width=.7bp,rounded corners=3bp]
  (414,90) rectangle (594,156);
\begin{scope}
  \clip[rounded corners=3bp] (414,90) rectangle (594,156);
  \fill[dpTeal!8] (414,90) rectangle (594,114);
\end{scope}
\node[anchor=west,inner sep=0,text=dpTeal,
  dpFont={15}{18}{12}]
  at (426,102) {\textbf{Camera trajectory}};
\node[anchor=west,inner sep=0,outer sep=0,text=dpInk,align=left,
  dpFont={14.5}{18}{12}]
  at (426,135) {\mbox{Trajectory annotation}\\\mbox{Trajectory filtering}};

\filldraw[fill=white,draw=dpPurple!45,line width=.7bp,rounded corners=3bp]
  (414,164) rectangle (594,245);
\begin{scope}
  \clip[rounded corners=3bp] (414,164) rectangle (594,245);
  \fill[dpPurple!8] (414,164) rectangle (594,188);
\end{scope}
\node[anchor=west,inner sep=0,text=dpRose,
  dpFont={15}{18}{12}]
  at (426,176) {\textbf{Video captioning}};
\node[anchor=west,inner sep=0,outer sep=0,text=dpInk,align=left,
  dpFont={14.5}{18}{12}]
  at (426,215.5) {\mbox{Semantic segmentation}\\\mbox{Multi-turn captioning}\\\mbox{Final review}};

\filldraw[fill=dpSampling!3,draw=dpSampling!35,line width=.7bp,rounded corners=3bp]
  (628,103) rectangle (826,232);
\begin{scope}
  \clip[rounded corners=3bp] (628,103) rectangle (826,232);
  \fill[dpSampling!9] (628,103) rectangle (826,135);
\end{scope}
\node[text=dpSampling,inner sep=0,dpFont={15}{18}{12}]
  at (727,119) {\textbf{Adaptive sampling}};
\node[text=dpMuted,inner sep=0,dpFont={14}{17}{10}]
  at (727,154) {Stage-wise distributions};
\filldraw[fill=white,draw=dpSampling!35,line width=.7bp,rounded corners=3bp]
  (638,175) rectangle (718,220);
\node[text=dpInk,inner sep=0,dpFont={15}{18}{12}]
  at (678,188) {Stage 1};
\node[text=dpMuted,inner sep=0,dpFont={14}{17}{10}]
  at (678,207) {Data mix 1};
\foreach \dotx in {724,727,730}{
  \fill[dpMuted] (\dotx,198) circle[radius=.6bp];}
\filldraw[fill=white,draw=dpSampling!35,line width=.7bp,rounded corners=3bp]
  (736,175) rectangle (816,220);
\node[text=dpInk,inner sep=0,dpFont={15}{18}{12}]
  at (776,188) {Stage 5};
\node[text=dpMuted,inner sep=0,dpFont={14}{17}{10}]
  at (776,207) {Data mix 5};

\pgfresetboundingbox
\path[use as bounding box] (0,0) rectangle (836,258);
\end{tikzpicture}%
\endgroup%
  }
  \caption{\textbf{Data pipeline.} Raw videos are split, transcoded, and filtered by basic video properties and visual quality. Hierarchical clustering supports within-cluster deduplication, and a VLM assigns labels to each video from a predefined taxonomy. Video tagging adds captions and, for selected clips, camera-trajectory annotations. Across each training stage, adaptive sampling adjusts the training distribution in response to weaknesses identified from evaluations of earlier stages.}
  \label{fig:data_pipeline_v2}
\end{figure*}
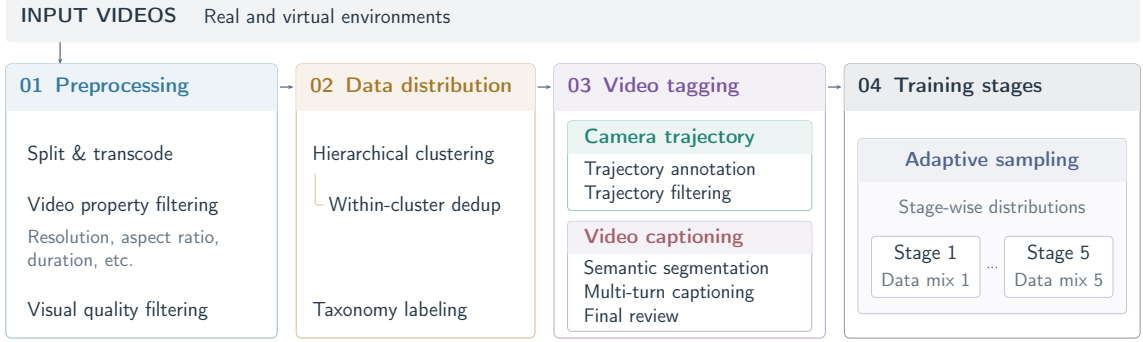

We construct a large-scale, heterogeneous video corpus spanning real and virtual environments. Our end-to-end pipeline converts raw videos into training samples through standardized preprocessing, filtering, semantic deduplication, camera-trajectory annotation, and structured captioning. We adjust the sampling distribution across training stages to address weaknesses identified in each stage. Figure \ref{fig:data_pipeline_v2} summarizes the pipeline.

\subsection{Data Filtering}
The raw videos vary in encoding types, duration, frame rate, resolution, and encoding quality. We standardize preprocessing by first dividing each video into 5-minute segments and then splitting these segments at detected shot boundaries. Our shot boundary detector is based on OmniShotCut \citep{wang2026omnishotcut} and fine-tuned on high-quality annotations to improve accuracy in challenging cases. We transcode the resulting clips into a unified format and filter them by basic video properties (e.g., duration, aspect ratio, and resolution) and visual quality metrics (e.g., motion, clarity, and aesthetic scores). Additional filters screen for derivative content and abrupt transitions.

\subsection{Data Distribution}
\paragraph{Clustering and deduplication}
We extract a visual embedding for each video using Qwen3-VL-Embedding \citep{qwen3vlembedding}, group videos by visual similarity through hierarchical clustering, and apply SemDeDup \citep{abbas2023semdedup} within each cluster to remove semantically redundant clips.

\paragraph{Taxonomy labeling}
We construct a hierarchical taxonomy covering semantic content and visual attributes. Content categories include different types of interaction, object manipulation, human and object movement, and camera motion; visual attributes describe camera perspective, subject type, subject and camera motion patterns, and recording methods. We fine tune a VLM to label each video according to this taxonomy, enabling targeted data selection and sampling along complementary semantic and visual dimensions.

\paragraph{Adaptive sampling}
We adapt the sampling distribution across training stages to address weaknesses identified in earlier stages. To select relevant data, we associate each video with two sets of labels: predicted video-level taxonomy labels and labels derived from hierarchical clustering. Since visual clustering is unsupervised, we manually annotate clusters at an intermediate level of the hierarchy to make them usable for data selection.

Depending on the target content category or visual attribute, we select videos using taxonomy labels, cluster-level annotations, or a combination of both. We then adjust the sampling weights of the selected subsets in subsequent training stages to prioritize areas that need further improvement.

\subsection{Camera Trajectories}
\paragraph{Trajectory annotation} 
We use ViPE \citep{huang2025vipe} to estimate camera trajectories. To manage its substantial computational cost, we apply it to a subset of videos suitable for trajectory estimation.

To scale annotation, we upgrade the camera calibration and segmentation modules with more capable pretrained models while retaining the existing depth estimator. We modularize the pipeline and apply adaptive spatiotemporal downsampling to reduce memory usage and improve throughput. We also tune the backend SLAM sliding window and optimization schedules to balance optimization scope against accumulated trajectory drift.

We additionally train a video-based inverse dynamics model (IDM), initialized from a pretrained VPT \citep{baker2022video} backbone, to predict discrete camera motion (such as WASD keyboard inputs) directly from video. Training combines synthetic camera-motion videos, and ViPE-derived motion labels refined using dense optical flow. Manually annotated shake and zoom data provide additional supervision, and unavailable labels are masked during training. The IDM predictions support camera motion categorization and data filtering, helping identify clips with inconsistent motion labels or unreliable trajectory annotations.

\paragraph{Trajectory filtering}
We filter in-the-wild videos both before ViPE estimation to reduce computational cost and afterward to exclude clips with static cameras, shot cuts, lens distortion, or local estimation instabilities. This filtering proceeds in four sequential steps, with only clips that pass each step advancing to the next:
\begin{enumerate}
[leftmargin=15pt]
  \item We restrict the optical-flow motion-score range to reject static scenes and abrupt spikes.
  \item For the remaining clips, we run CoTracker3~\citep{karaev2025cotracker3} to track a uniform grid of points and filter out videos in which the inter-frame displacement of visible points exceeds $20\%$ of $\max(H,W)$ or the median total displacement of tracked points is below $30\%$ of $
  \max(H,W)$.
  \item We then use AnyCalib~\citep{tiradogarin2025anycalib} to jointly estimate per-frame focal length and radial distortion, penalizing clips with excessive focal-length variation, abrupt focal-length changes, overly large focal lengths, or strong distortion.
  \item Finally, retained clips undergo ViPE estimation, followed by a trajectory audit to detect optimization failures and degenerate motion scales. We reject trajectories exhibiting repeated large orientation jumps within local windows, excessive inter-frame viewing-angle offsets, translation
  spikes relative to the local mean inter-frame displacement, or insufficient overall camera displacement.
\end{enumerate}

\subsection{Video Captioning}
\begin{figure*}[t]
  \centering
  \ifdefined\captionsetup
    \captionsetup{skip=4pt}
  \else
    \setlength{\abovecaptionskip}{4pt}
  \fi
  \resizebox{\linewidth}{!}{%
\begingroup%
\definecolor{cpInk}{HTML}{263746}%
\definecolor{cpMuted}{HTML}{667583}%
\definecolor{cpBlue}{HTML}{347AA0}%
\definecolor{cpAmber}{HTML}{A57829}%
\definecolor{cpPurple}{HTML}{7955A0}%
\definecolor{cpTeal}{HTML}{238476}%
\definecolor{cpLine}{HTML}{D6DFE5}%
\newcommand{\cpText}[5]{%
  \node[anchor=north west,inner sep=0,outer sep=0,text=#4,align=left,%
    cpFont={#3}{20}{12}]%
    at (#1,#2) {#5};}%
\newcommand{\cpCard}[6]{%
  \filldraw[fill=white,draw=#5!50,line width=.8bp,rounded corners=4bp]%
    (#1,#2) rectangle (#3,#4);%
  \begin{scope}%
    \clip[rounded corners=4bp] (#1,#2) rectangle (#3,#4);%
    \fill[#5!9] (#1,#2) rectangle (#3,{#2+36});%
  \end{scope}%
  \node[anchor=west,inner sep=0,text=#5,%
    cpFont={17}{20}{12}]%
    at ({#1+12},{#2+18}) {\strut\textbf{#6}};}%
\begin{tikzpicture}[x=1bp,y=-1bp,
  cpFont/.style n args={3}{
    font={\pgfmathsetmacro{\cpLeading}{#2*#3/#1}%
      \sffamily\fontsize{#3}{\cpLeading}\selectfont},
    scale={#1/#3}},
  flow/.style={draw=cpMuted,line width=.9bp,
    -{Stealth[length=4bp,width=3bp]},rounded corners=3bp}]

\cpCard{0}{0}{188}{266}{cpAmber}{Video segmentation}
\cpCard{200}{0}{432}{266}{cpBlue}{First turn}
\cpCard{442}{0}{684}{266}{cpPurple}{Subsequent turns}
\cpCard{694}{0}{1060}{266}{cpTeal}{Final turn}
\foreach \a/\b in {190/200,432/442,684/694}{
  \draw[flow,shorten <=2bp,shorten >=2bp] (\a,18)--(\b,18);}

\cpText{14}{61}{16}{cpInk}{Visual features}
\cpText{14}{94}{15.5}{cpInk}{Find split points}
\draw[cpAmber,line width=1.1bp,line join=round]
  (14,155)--(30,154)--(42,156)--(52,134)--(60,155)
  --(98,155)--(108,134)--(116,155)--(176,154);
\foreach \x in {52,108}{
  \draw[cpAmber!65,densely dashed,line width=.65bp] (\x,134)--(\x,211);}
\foreach \a/\b/\lab in {14/52/1,52/108/2,108/176/N}{
  \filldraw[fill=cpAmber!12,draw=cpAmber!60,line width=.6bp]
    (\a,181) rectangle (\b,211);
  \node[text=cpAmber,inner sep=0,cpFont={15.5}{19}{12}]
    at ({(\a+\b)/2},196) {$s_{\lab}$};}
\cpText{14}{238}{15}{cpMuted}{Fixed timestamps}

\filldraw[fill=cpBlue!9,draw=cpBlue!45,line width=.65bp,rounded corners=2bp]
  (214,48) rectangle (418,80);
\node[text=cpBlue,inner sep=0,cpFont={16}{19}{12}]
  at (316,64) {Video segment $s_1$};
\cpText{214}{98}{15.5}{cpInk}{Initial scene \& camera}
\cpText{214}{128}{15.5}{cpInk}{Subject IDs \& static profiles}
\cpText{214}{158}{15.5}{cpInk}{Perspective \& visual style}
\cpText{214}{188}{15.5}{cpInk}{Timestamped subject actions}
\fill[cpBlue!7,rounded corners=2bp] (214,222) rectangle (418,254);
\node[text=cpBlue,inner sep=0,cpFont={16}{19}{12}]
  at (316,238) {\strut\textbf{English segment draft}};

\filldraw[fill=cpPurple!9,draw=cpPurple!45,line width=.65bp,rounded corners=2bp]
  (456,48) rectangle (670,80);
\node[text=cpPurple,inner sep=0,cpFont={16}{19}{12}]
  at (563,64) {Video segment $s_k$};
\cpText{456}{98}{15.5}{cpInk}{Changes vs. previous segment}
\cpText{456}{128}{15.5}{cpInk}{Scene/camera events in order}
\cpText{456}{158}{15.5}{cpInk}{Update subjects \& visual style}
\cpText{456}{188}{15.5}{cpInk}{Timestamped subject actions}
\fill[cpPurple!7,rounded corners=2bp] (456,222) rectangle (670,254);
\node[text=cpPurple,inner sep=0,cpFont={16}{19}{12}]
  at (563,238) {\strut\textbf{English segment draft}};

\fill[cpTeal!8,rounded corners=2bp] (708,48) rectangle (1046,80);
\node[text=cpTeal,inner sep=0,cpFont={15.5}{19}{12}]
  at (877,64) {Text only / no new video tokens};
\cpText{708}{98}{16}{cpTeal}{\textbf{Review \& translate}}
\cpText{879}{98}{16}{cpTeal}{\textbf{Validation}}
\draw[cpLine,line width=.6bp] (867,98)--(867,209);
\cpText{708}{132}{15}{cpInk}{Merge \& revise\\annotations}
\cpText{708}{176}{15}{cpInk}{Translate EN to ZH}
\cpText{879}{132}{15}{cpInk}{JSON structure \& fields,\\tags \& subject refs}
\cpText{879}{176}{15}{cpInk}{Segment/action times}
\cpText{879}{203}{15}{cpInk}{EN/ZH consistency}
\fill[cpTeal!9,rounded corners=2bp] (708,222) rectangle (1046,254);
\node[text=cpTeal,inner sep=0,cpFont={16}{19}{12}]
  at (877,238) {\strut\textbf{Final EN / ZH JSON}};

\draw[cpMuted!65,line width=.75bp]
  (200,274)--(200,281)--(684,281)--(684,274);
\node[text=cpMuted,inner sep=0,cpFont={15}{18}{12}]
  at (442,297) {Context: current + prior segments and annotations};

\pgfresetboundingbox
\path[use as bounding box] (0,0) rectangle (1060,306);
\end{tikzpicture}%
\endgroup%
  }
  \caption{\textbf{Structured video captioning.} The VLM annotates segments sequentially, maintains subject identities, interleaves scene and camera descriptions chronologically, and records subject actions separately with timestamps. A final text-only turn reviews, translates, and validates the annotations to produce aligned English and Chinese captions. During training, camera-motion sentences and their tags are removed when camera trajectories are available and retained otherwise.}
  \label{fig:caption_json}
\end{figure*}
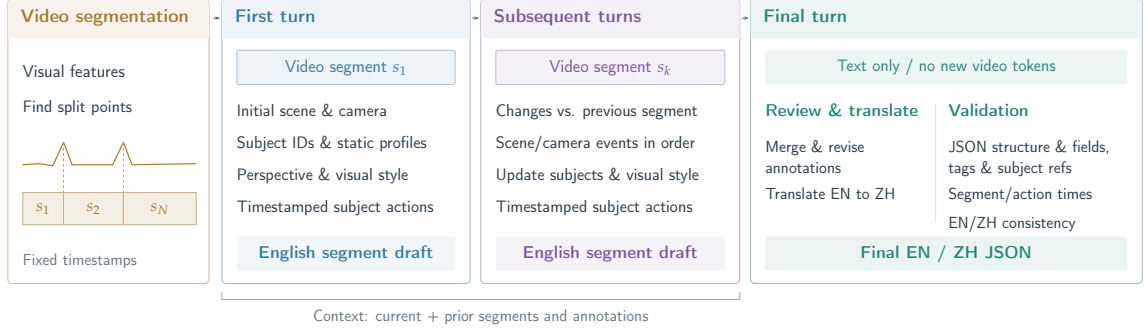

We design structured video captions to preserve subject identity, scene context, and action chronology, with camera motion explicitly tagged for flexible supervision. Figure \ref{fig:caption_json} and \ref{fig:caption_schema} illustrates the caption design and multi-turn annotation procedure.


\paragraph{Structured caption}
Our captions use structured JSON with video-level and segment-level information. Video-level fields specify visual style, camera perspective, and a shared subject list. Each subject has an explicit identifier and a static profile anchored to its earliest clearly identifiable appearance; subsequent actions and state changes reference this identifier to maintain continuity across segments. Within each segment, scene descriptions and camera-motion sentences are interleaved chronologically in the description field. Subject actions and their visible outcomes are recorded separately, with start and end timestamps for each action.

The first segment establishes the scene; subsequent segments describe changes relative to the preceding segment without repeating unchanged context. Action and camera-motion descriptions are limited to each segment's temporal boundaries, and action entries form a chronological, non-overlapping sequence. This representation records scene changes and action timing to support segment-wise autoregressive rollout training.

\paragraph{Camera-motion description}
We enclose camera-motion sentences in explicit \textless{}CameraN\textgreater{} and \textless{}/CameraN\textgreater{} tags so they can be selected independently of scene and action descriptions. For training segments with ViPE-estimated trajectories, we remove the camera-motion descriptions and their enclosing tags, using the estimated trajectories for explicit camera-motion supervision. For segments without trajectory annotations, we retain the tagged sentences as textual supervision. This complementary strategy supports camera control through either keyboard inputs mapped to trajectories or natural-language motion instructions.

\paragraph{Multi-turn annotation}
We generate captions in two stages. First, we extract visual features using V-JEPA 2 \citep{assran2025vjepa2} and divide each clip into shorter segments based on temporal changes in these features. In the second stage, a VLM annotates the segments sequentially in a multi-turn conversation.

The first turn establishes the initial scene and subject profiles; subsequent turns describe scene changes and current actions while refining subject identities and video-level labels using accumulated evidence. Previous segments and annotations remain in the conversation context to support temporal and semantic consistency.

For subsequent segments, the VLM uses visual evidence only from the current and preceding segments to ground its descriptions and reduce hallucination. After all segments have been processed, a final text-only turn reviews the accumulated annotations, corrects earlier drafts, and resolves inconsistencies while preserving the supplied segment boundaries. It produces the final English caption and a structurally aligned Chinese translation, both following the format in Figure \ref{fig:caption_schema}. In this same turn, the VLM checks the JSON structure, required fields, subject references, tags, segment and action timestamps, and consistency between different language versions, then corrects detected inconsistencies before finalizing the output.

\begin{figure*}[t]
\centering
\begin{minipage}{0.97\textwidth}
\begin{lstlisting}[style=captionjson]
{
  "perspective_view": "[Video-level perspective]",
  "Style": "[Visual style category]",
  "subject_list": [
    {
      "subject_id": "<|Subject1|>",
      "display_name": "[Concise subject name]",
      "subject_attribute": "[Static profile at the earliest clear appearance]"
    }
  ],
  "segments": [
    {
      "start_time": "[Segment 1 start: MM:SS.mmm]",
      "end_time": "[Segment 1 end: MM:SS.mmm]",
      "visual_description_long": "A detailed description of the setting, background, and objects visible in the first segment. <Camera1>A description of the camera's movement and the resulting framing changes</Camera1>.",
      "description_change": null,
      "contained_actions": [
        {
          "start_time": "[Action 1 start: MM:SS.mmm]",
          "end_time": "[Action 1 end: MM:SS.mmm]",
          "action_description": "<Action1><|Subject1|> [first action and visible outcome within the current segment]</Action1>."
        },
        {
          "start_time": "[Action 2 start: MM:SS.mmm]",
          "end_time": "[Action 2 end: MM:SS.mmm]",
          "action_description": "<Action2><|Subject1|> [subsequent action and visible outcome]</Action2>."
        }
      ]
    },
    {
      "start_time": "[Segment 2 start: MM:SS.mmm]",
      "end_time": "[Segment 2 end: MM:SS.mmm]",
      "visual_description_long": null,
      "description_change": "Compared to the previous segment, ... (summarize changes in the setting, background, and objects). <Camera1>A description of the camera's movement within the current segment</Camera1>.",
      "contained_actions": []  // Can be empty if no subject action is visible in this segment.
    },
    {
      "start_time": "[Segment 3 start: MM:SS.mmm]",
      "end_time": "[Segment 3 end: MM:SS.mmm]",
      "visual_description_long": null,
      "description_change": "Compared to the previous segment, ... (summarize changes in the setting, background, and objects). <Camera1>Camera movement at this moment</Camera1>. The next scene or object change. <Camera2>A subsequent change in camera movement</Camera2>. Further scene or object change.",
      "contained_actions": [
        {
          "start_time": "[Action 1 start: MM:SS.mmm]",
          "end_time": "[Action 1 end: MM:SS.mmm]",
          "action_description": "<Action1><|Subject1|> [first action and visible outcome in this segment]</Action1>."
        }
      ]
    }
  ]
}
\end{lstlisting}
\end{minipage}
\caption{\textbf{Structured caption format}. Video-level fields define perspective, style, and shared subject profiles. Segment descriptions interleave scene information and camera-tagged motion chronologically, excluding main-subject references. The first segment provides a full description; later segments describe changes. Subject actions and their visible outcomes are recorded separately in action-tagged entries using subject tokens. Action intervals are non-overlapping and remain within segment boundaries.}
\label{fig:caption_schema}
\end{figure*}

\paragraph{Caption models}
We use different captioning models across training stages to balance annotation cost and caption quality. For data-intensive stages, we employ Gemma 4~\citep{gemmateam2026gemma4} and Qwen3.5-122B-A10B~\citep{qwen35blog} for efficient large-scale captioning. For the annealing stage, where annotation quality is prioritized over throughput, we use Kimi K3~\citep{moonshot2026kimik3} to generate detailed segment-wise annotations.

\section{Model Architecture}
\label{sec:model_architecture}
\begin{figure}[!t]
  \centering
  \includegraphics[width=\textwidth]{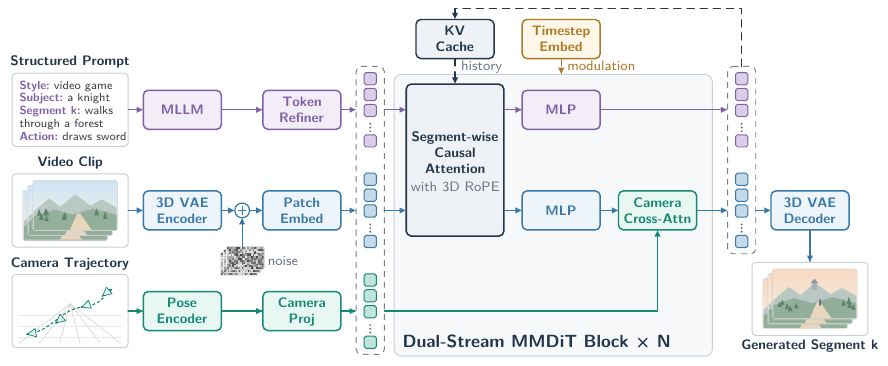}
  \caption{\textbf{Overview of the model architecture.}
A structured multi-segment prompt, a video clip, and a camera trajectory are encoded into text, video, and camera tokens, respectively.
Text and video tokens are processed by $N=54$ dual-stream MMDiT blocks.
Within each block, segment-wise causal attention with 3D RoPE allows tokens to attend to the current and preceding segments, while diffusion timestep embeddings modulate both streams.
The two streams pass through separate MLPs, and camera cross-attention further conditions the video stream on camera tokens.
During autoregressive generation, keys and values associated with both text and video tokens are stored in the KV cache as context for subsequent segments.
The 3D VAE decoder maps denoised video latents to RGB frames.
Purple, blue, and green denote the text, video, and camera streams, respectively.}
  \label{fig:overview}
\end{figure}

\shortname builds upon HunyuanVideo 1.5 \citep{wu2025hunyuanvideo}. We introduce segment-wise causal attention for autoregressive world generation (Section \ref{subsec:causal_masking}) and camera cross-attention for explicit camera control (Section \ref{subsec:camera_conditioning}). \shortname uses a DCAE-V video autoencoder \citep{chen2025dc}\footnote{We thank the NVIDIA DC-AE team for generously sharing their model checkpoints and providing valuable support in integrating DC-AE-V into HunyuanVideo 1.5.} and a frozen Qwen2.5-VL-7B text encoder \citep{bai2025qwen25vl} inherited from HunyuanVideo 1.5, followed by a timestep-aware token refiner. Starting from Gaussian noise, the model denoises video latents one segment at a time, conditioned on the current text prompt, an optional camera trajectory, and cached keys and values from preceding segments. A dedicated cross-attention module in each MMDiT block~\citep{esser2024scaling} injects camera conditioning, using video features as queries and frame-aligned camera tokens as keys and values. Figure \ref{fig:overview} illustrates the architecture, and Table \ref{tab:architecture_settings} summarizes its primary configuration.

\begin{table*}[!t]
    \centering
    \small
    \setlength{\tabcolsep}{5pt}
    \begin{tabular}{@{}ll@{}}
        \toprule
        Component & Configuration \\
        \midrule
        Video autoencoder & DCAE-V, $32\times$ spatial and $4\times$ temporal compression \\
        Latent representation & 32 channels; patch size $1\times1\times1$ \\
        Text encoder & Qwen2.5-VL-7B, width 3,584, maximum length 1,280 \\
        Transformer backbone & 54 dual-stream MMDiT blocks \\
        Hidden width / MLP width & 2,048 / 8,192 \\
        Self-attention & 16 heads, 128 dimensions per head \\
        Video RoPE allocation & $(16,56,56)$ for time, height, and width \\
        Camera representation & 64-D grouped camera extrinsics \\
        Camera cross-attention & width 1,024; 16 heads; one module per MMDiT block \\
        Prediction target & 32-channel latent flow velocity \\
        \bottomrule
    \end{tabular}
    \caption{Summary of primary architectural settings.}
    \vspace{-3mm}
\label{tab:architecture_settings}
\end{table*}

\paragraph{Latent video representation}
\label{subsec:latent_video}
We encode an RGB video $X \in \mathbb{R}^{B \times 3 \times F \times H \times W}$ into a compact spatiotemporal representation using a causal DCAE-V encoder $\mathcal{E}$:
\begin{equation}
    Z = \mathcal{E}(X)
    \in \mathbb{R}^{B \times 32 \times F_z \times H_z \times W_z},
    \qquad
    F_z = \left\lfloor\frac{F-1}{4}\right\rfloor+1,
    H_z=\left\lfloor\frac{H}{32}\right\rfloor, W_z=\left\lfloor\frac{W}{32}\right\rfloor
    \label{eq:vae_encoding}
\end{equation}
with spatial and temporal compression factors of $32$ and $4$, respectively. Here, $F$ denotes the number of frames, while $H$ and $W$ denote the height and width of each frame, respectively. Each latent position is treated as an individual token ($1 \times 1 \times 1$ patch) and linearly projected to the DiT hidden dimension of $2048$. The diffusion transformer operates directly in this latent space, with the resulting denoised representation decoded back into RGB video by a decoder $\mathcal{D}$.

\paragraph{Structured text conditioning}
During preprocessing, each video segment is annotated with a structured JSON caption, as illustrated in Figure~\ref{fig:caption_json}. A frozen Qwen2.5-VL-7B encoder~\citep{bai2025qwen25vl} extracts segment-wise text embeddings, which are subsequently processed by a two-layer, timestep-conditioned token refiner. Specifically, the embeddings are projected from 3,584 to the MMDiT hidden dimension of 2,048, followed by adaptive modulation conditioned on the diffusion timestep and pooled text context. The resulting refined embeddings serve as text tokens for attention in the MMDiT backbone.

\begin{figure*}[t]
  \centering

  \definecolor{dpInk}{HTML}{263746}
  \definecolor{dpBlue}{HTML}{347AA0}
  \definecolor{dpAmber}{HTML}{A57829}
  \definecolor{dpPurple}{HTML}{7955A0}
  \definecolor{dpTeal}{HTML}{238476}
  \definecolor{dpLine}{HTML}{D6DFE5}

  \colorlet{vv}{dpBlue!40!white}
  \colorlet{vt}{dpTeal!35!white}
  \colorlet{tv}{dpAmber!40!white}
  \colorlet{tt}{dpPurple!35!white}

  \colorlet{masked}{dpInk!4!white}
  \colorlet{minorgrid}{dpLine}
  \colorlet{majorgrid}{dpInk!65!white}

  \colorlet{videohead}{dpBlue!9!white}
  \colorlet{texthead}{dpPurple!9!white}

  \newcommand{\DrawAttentionMask}[1]{%
    \begin{tikzpicture}[
      x=0.62cm,
      y=0.62cm,
      font=\fontfamily{phv}\selectfont\mdseries,
      text=dpInk,
      minor/.style={draw=minorgrid, line width=0.35pt},
      major/.style={draw=majorgrid, line width=1.05pt},
      token/.style={
        font=\fontfamily{phv}\selectfont\mdseries\small
      },
      grouplabel/.style={
        font=\fontfamily{phv}\selectfont\mdseries\small
      }
    ]
      \path[use as bounding box]
        (-3.35,3.05) rectangle (15.35,-15.35);

      \path[fill=masked] (0,0) rectangle (15,-15);

      \ifnum#1=0\relax

        \path[fill=vv] (0,0) rectangle (3,-3);
        \path[fill=vv] (0,-3) rectangle (6,-6);

        \path[fill=vt] (6,0) rectangle (10.5,-3);
        \path[fill=vt] (6,-3) rectangle (15,-6);

        \path[fill=tv] (0,-6) rectangle (3,-10.5);
        \path[fill=tv] (0,-10.5) rectangle (6,-15);

        \path[fill=tt] (6,-6) rectangle (10.5,-10.5);
        \path[fill=tt] (6,-10.5) rectangle (15,-15);

      \else

        \path[fill=vv] (0,0) rectangle (3,-3);
        \path[fill=vv] (0,-3) rectangle (6,-6);

        \path[fill=vt] (6,0) rectangle (7.125,-3);
        \path[fill=vt] (7.125,-0.6) rectangle (8.25,-1.2);
        \path[fill=vt] (8.25,-1.8) rectangle (10.5,-2.4);
        \path[fill=vt] (6,-3) rectangle (7.125,-6);
        \path[fill=vt] (10.5,-3) rectangle (11.625,-6);
        \path[fill=vt] (11.625,-3.6) rectangle (12.75,-4.2);
        \path[fill=vt] (12.75,-4.8) rectangle (15,-5.4);

        \path[fill=tv] (0,-6) rectangle (3,-10.5);
        \path[fill=tv] (0,-10.5) rectangle (6,-15);

        \path[fill=tt] (6,-6) rectangle (7.125,-10.5);
        \path[fill=tt] (7.125,-7.125) rectangle (8.25,-8.25);
        \path[fill=tt] (8.25,-8.25) rectangle (10.5,-10.5);
        \path[fill=tt] (6,-10.5) rectangle (7.125,-15);
        \path[fill=tt] (10.5,-10.5) rectangle (11.625,-15);
        \path[fill=tt] (11.625,-11.625) rectangle (12.75,-12.75);
        \path[fill=tt] (12.75,-12.75) rectangle (15,-15);
      \fi

      \foreach \x in {0.6,1.2,1.8,2.4,3.6,4.2,4.8,5.4}{
        \draw[minor] (\x,0) -- (\x,-15);
      }
      \foreach \y in {
        -0.6,-1.2,-1.8,-2.4,-3.6,-4.2,-4.8,-5.4
      }{
        \draw[minor] (0,\y) -- (15,\y);
      }

      \foreach \x in {7.125,8.25,11.625,12.75}{
        \draw[minor] (\x,0) -- (\x,-15);
      }
      \foreach \y in {-7.125,-8.25,-11.625,-12.75}{
        \draw[minor] (0,\y) -- (15,\y);
      }

      \draw[major] (0,0) rectangle (15,-15);
      \foreach \x in {3,6,10.5}{
        \draw[major] (\x,0) -- (\x,-15);
      }
      \foreach \y in {-3,-6,-10.5}{
        \draw[major] (0,\y) -- (15,\y);
      }

      \path[fill=videohead] (0,1.05) rectangle (3,1.60);
      \path[fill=videohead] (3,1.05) rectangle (6,1.60);
      \path[fill=texthead] (6,1.05) rectangle (10.5,1.60);
      \path[fill=texthead] (10.5,1.05) rectangle (15,1.60);

      \node[grouplabel] at (1.5,1.325) {Video $s_0$};
      \node[grouplabel] at (4.5,1.325) {Video $s_1$};
      \node[grouplabel] at (8.25,1.325) {Text $s_0$};
      \node[grouplabel] at (12.75,1.325) {Text $s_1$};

      \node[
        font=\fontfamily{phv}\selectfont\bfseries\Large,
        anchor=center
      ] at (7.5,2.45) {Key / value tokens};

      \foreach \x/\lab in {
        0.3/{},0.9/{$a_0$},1.5/{},2.1/{$a_1$},2.7/{},
        3.3/{},3.9/{$a_0$},4.5/{},5.1/{$a_1$},5.7/{},
        6.5625/{head},7.6875/{$a_0$},9.375/{$a_1$},
        11.0625/{head},12.1875/{$a_0$},13.875/{$a_1$}
      }{
        \node[token,rotate=45,anchor=west]
          at (\x,0.16) {\lab};
      }

      \path[fill=videohead] (-1.95,0) rectangle (-1.40,-3);
      \path[fill=videohead] (-1.95,-3) rectangle (-1.40,-6);
      \path[fill=texthead] (-1.95,-6) rectangle (-1.40,-10.5);
      \path[fill=texthead] (-1.95,-10.5) rectangle (-1.40,-15);

      \node[grouplabel,rotate=90]
        at (-1.675,-1.5) {Video $s_0$};
      \node[grouplabel,rotate=90]
        at (-1.675,-4.5) {Video $s_1$};
      \node[grouplabel,rotate=90]
        at (-1.675,-8.25) {Text $s_0$};
      \node[grouplabel,rotate=90]
        at (-1.675,-12.75) {Text $s_1$};

      \node[
        font=\fontfamily{phv}\selectfont\bfseries\Large,
        rotate=90,
        anchor=center
      ] at (-2.75,-7.5) {Query tokens};

      \foreach \y/\lab in {
        -0.3/{},-0.9/{$a_0$},-1.5/{},-2.1/{$a_1$},-2.7/{},
        -3.3/{},-3.9/{$a_0$},-4.5/{},-5.1/{$a_1$},-5.7/{},
        -6.5625/{head},-7.6875/{$a_0$},-9.375/{$a_1$},
        -11.0625/{head},-12.1875/{$a_0$},-13.875/{$a_1$}
      }{
        \node[token,anchor=east] at (-0.05,\y) {\lab};
      }
    \end{tikzpicture}%
  }

  \begin{tikzpicture}[text=dpInk]
    \node[
      inner sep=0,
      outer sep=0,
      anchor=north west
    ] (left) at (0,0) {%
      \resizebox{0.455\textwidth}{!}{\DrawAttentionMask{0}}%
    };

    \node[
      inner sep=0,
      outer sep=0,
      anchor=north west
    ] (right) at (0.545\textwidth,0) {%
      \resizebox{0.455\textwidth}{!}{\DrawAttentionMask{1}}%
    };

    \node[
      anchor=north,
      font=\fontfamily{phv}\selectfont\mdseries\small
    ] at ($(left.south)+(0,-2mm)$)
      {(a) segment-wise causal};

    \node[
      anchor=north,
      font=\fontfamily{phv}\selectfont\mdseries\small
    ] at ($(right.south)+(0,-2mm)$)
      {(b) action-aware causal};
  \end{tikzpicture}

  \par\vspace{2mm}

  \begin{tikzpicture}[
    font=\fontfamily{phv}\selectfont\mdseries\scriptsize,
    text=dpInk,
    legendtext/.style={anchor=west,inner sep=0pt}
  ]
    \foreach \x/\col/\lab in {
      0/vv/{video $\to$ video},
      2.65/vt/{video $\to$ text},
      5.20/tv/{text $\to$ video},
      7.75/tt/{text $\to$ text}
    }{
      \path[fill=\col] (\x,0) rectangle ++(0.27,0.18);
      \node[legendtext] at (\x+0.38,0.09) {\lab};
    }

    \path[fill=masked,draw=minorgrid]
      (10.25,0) rectangle ++(0.27,0.18);
    \node[legendtext] at (10.63,0.09) {masked};
  \end{tikzpicture}

  \caption{\textbf{Comparison of segment-wise and action-aware causal attention.}
    (a) Tokens attend to all video and text tokens in the current
    and preceding segments.
    (b) Action-aware masking further restricts access to action-specific
    text tokens.
    Colors indicate query-to-key modality pairs; gray cells are masked.
    \textit{head} denotes the environment description for the current segment,
    and $a_0,\ldots,a_n$ denote actions within that segment.}
  \label{fig:attention_mask_comparison}
\end{figure*}
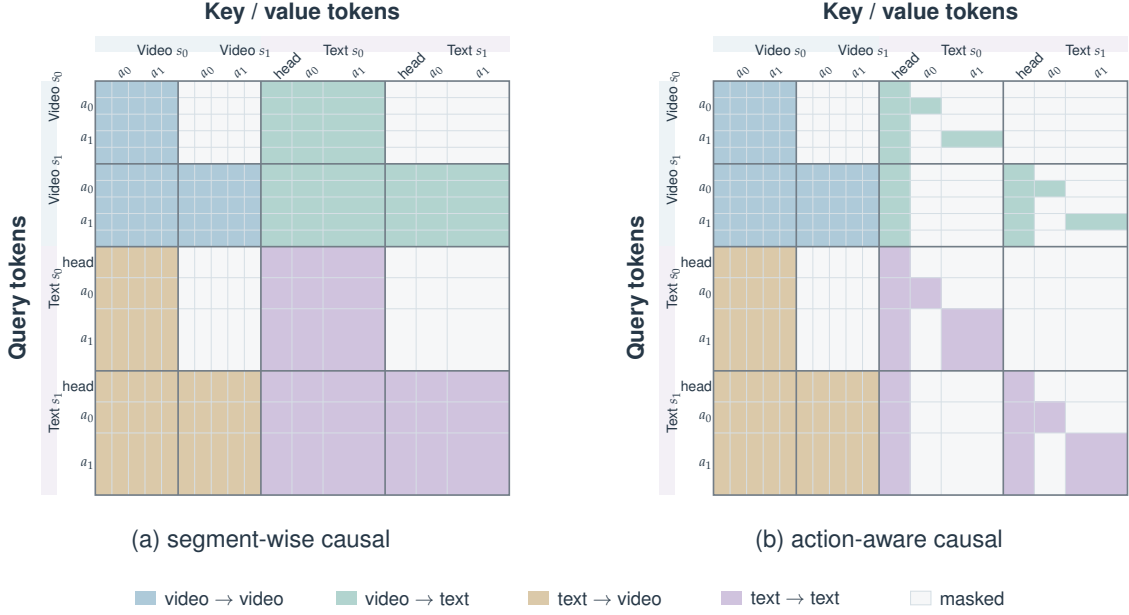

\paragraph{Flow-matching objective}
\label{subsec:flow_matching}
The model is trained with linear flow matching~\citep{lipman2023flow}. Given a clean video latent $Z$, Gaussian noise $\epsilon$, and $t\in[0,1]$, we construct
\begin{equation}
    Z_t = (1-t)Z + t\epsilon,
    \qquad
    V_t = \epsilon-Z.
    \label{eq:flow_path}
\end{equation}
The network predicts the velocity $V_t$ along this path. At inference time, the reverse ODE is initialized from Gaussian noise at $t=1$ and integrated toward $t=0$ using 100 Euler steps. Classifier-free text guidance~\citep{ho2022classifierfree} and camera guidance are applied during integration. The resulting clean 32-channel latent is finally decoded by $\mathcal{D}$ to obtain the RGB video.

\subsection{Segment-wise Causal Attention}
\label{subsec:causal_masking}
Unlike conventional video generation, video-based world models simulate evolving world states in temporal order, requiring a causal modeling framework. However, the fully bidirectional attention in our base text-to-video (T2V) model does not enforce this causal structure. We therefore replace it with segment-wise causal attention aligned with the temporal structure of the input captions. Specifically, each caption includes explicit temporal boundaries that partition the video and text tokens into temporally aligned semantic segments. Attention is bidirectional within each segment and causal across segments, allowing tokens to attend only to the current and preceding segments. We illustrate this design in Figure~\ref{fig:attention_mask_comparison}(a). Query tokens in segment $j$ can attend to all key and value tokens from the current and preceding segments $0,\ldots,j$, but not to future segments, enforcing causal information flow when modeling evolving world states.

We adapt the bidirectional attention to segment-wise causal attention during the early training stages. However, we observe two limitations of this masking design:
\begin{enumerate}
[leftmargin=15pt]
   \item \textbf{Within-segment action misalignment.} We observe that actions within a segment can occur in the wrong temporal order. This behavior stems from unrestricted attention to text tokens within each segment: query tokens can attend to the entire segment caption, including actions happening from a different time. This weakens the temporal alignment between video tokens and their corresponding action descriptions in the same segment.
    \item \textbf{Cross-segment action repetition.} We also observe that previously generated actions can recur across segments as generation progresses. This behavior is rooted in the continued accessibility of past action descriptions: video tokens in later segments can still attend to action descriptions from earlier segments, prompting the model to replay past actions.
\end{enumerate}

To address these issues, we introduce \textbf{action-aware causal attention}, which extends segment-wise causal attention with a finer-grained masking scheme to distinguish actions within and across segments, as illustrated in Figure~\ref{fig:attention_mask_comparison}(b). Each segment's text tokens are divided into a descriptive \emph{head} for the segment's environment and action \emph{tails} $a_0,\ldots,a_n$, each specifying an individual action and its corresponding temporal window.
This masking scheme addresses the two aforementioned issues as follows:
\begin{enumerate}
[leftmargin=15pt]
    \item \textbf{Within-segment action alignment.} Query tokens within an action window are restricted to attending only to the associated text tokens and the descriptive head tokens from the same segment. This preserves environmental context while reducing action-order errors and premature action execution.
    
\item \textbf{Cross-segment action isolation.} Descriptive heads remain accessible to the current and subsequent segments, whereas action tails are restricted to their originating segment. This preserves environmental context across segments while blocking access to past action descriptions, reducing action repetition in later segments.
\end{enumerate}

Applying action-aware causal attention during the later training stages substantially reduces action-order errors and cross-segment action repetition.

\subsection{Camera-Trajectory Conditioning}
\label{subsec:camera_conditioning}
\paragraph{Camera trajectory encoding and injection.}
\shortname{} supports continuous camera control during
segment-wise autoregressive generation. For each segment,
given camera-to-world poses $P_t$, we encode frame-to-frame
relative motion as $\Delta P_t=P_{t-1}^{-1}P_t$,
with the first increment set to the identity.
Relative transformations are computed independently within
each segment, avoiding dependence on the global coordinate
frame and camera poses from preceding segments.

Frame-to-frame translations are normalized by a single video-level scale, defined as the 99th percentile of translation magnitudes across all frames. This limits the influence of extreme values on the normalization scale while preserving relative motion speeds. We pad the beginning of the camera sequence by repeating its first transformation until the sequence length is divisible by four. Each group of four consecutive $4\times4$ transformation matrices is then flattened into a 64-dimensional vector and processed by a linear projection followed by a residual MLP, yielding one camera token per VAE latent frame. In each MMDiT block, visual tokens attend to temporally aligned camera tokens within the same segment through cross-attention:
\begin{equation}
    \widetilde V = V + \operatorname{CrossAttn}(V, C).
\end{equation}
Here $V$ denotes the visual tokens, and $C$ denotes the temporally aligned camera tokens within the same segment. In cross-attention, visual tokens serve as queries, while camera tokens provide keys and values. Recent interactive world models instead rely on coarser action interfaces: some quantize both camera translation and rotation into fixed directional commands~\citep{mao2025yume,happyoyster2026}, while others keep rotation continuous but restrict translation to discrete keyboard actions~\citep{yu2025gamefactory,he2026matrixgame2,robbyant2026lingbot}. Our relative-pose trajectory keeps both translation and rotation continuous, enabling finer-grained control over direction, speed, and scale.

\paragraph{Segment-wise mixed-control training.}
To support interleaved camera and text control during inference, we randomly disable camera conditioning on a per-segment basis during training, exposing the model to alternation between camera- and text-controlled segments. For segments with camera conditioning enabled, we remove camera descriptions from video caption to prevent the model from using them as shortcuts during training. For segments without camera conditioning, we retain these descriptions to enable text-driven camera control.

\paragraph{From keyboard inputs to camera trajectories.}
At inference, keyboard inputs are converted into camera trajectories. \texttt{W}/\texttt{S} control forward and backward motion, \texttt{A}/\texttt{D} control lateral motion, and mouse or arrow-key inputs control yaw and pitch rotations. These translation and rotation increments are accumulated frame by frame to construct camera-to-world poses. Motion speed, scale and duration can be adjusted without retraining.

\section{Training Infrastructure}
\label{sec:training_infra}

To efficiently process long, variable-length videos when adapting a pretrained video generator into an interactive world model through multi-stage training, we optimize distributed parallelism, data loading, and memory usage. This section describes the detailed infrastructure optimizations.

\subsection{Model and Data Parallelism}
\label{subsec:parallelism_optimization}

Training long, token-packed video sequences requires distributing both model state and activation computation across multiple GPUs. We combine Ulysses-style sequence parallelism (SP)~\citep{jacobs2023deepspeed}, data parallelism (DP), and FSDP2 hybrid sharding~\citep{pytorch_fsdp2}. Sequence parallelism partitions a single packed sample across a small group of GPUs, data parallelism assigns different packed samples to different groups, and FSDP2 shards model parameters, gradients, and optimizer states within each group. This organization enables sequences of up to 360K tokens while keeping parameter-sharding and sequence-parallel communication localized to small groups. On top of this parallel layout, we use a mask-flexible attention kernel for the heterogeneous segment- and action-aware masks, FLOPs-balanced sequence packing, a distributed Muon optimizer, and asynchronous checkpointing, which together determine the end-to-end throughput and robustness of long-running jobs.
 
\paragraph{Sequence parallelism.}
We adopt Ulysses sequence parallelism~\citep{jacobs2023deepspeed} with $N_{\mathrm{SP}}=8$ to reduce activation memory for packed video and text sequences of up to 360K tokens. Both video and text tokens are partitioned along the sequence dimension across eight GPUs, with each rank holding approximately 45K tokens at maximum sequence length. During attention, an all-to-all collective transforms sequence shards into head shards, allowing each rank to compute a subset of attention heads over the full sequence, followed by a reverse all-to-all to restore sequence sharding for local normalization, residual, and MLP operations.

\paragraph{Data parallelism.}
Each eight-GPU SP group forms one logical data-parallel replica and processes a different packed batch. For $N_{\mathrm{GPU}}$ total GPUs and sequence-parallel degree $N_{\mathrm{SP}}$, the number of replicas is $N_{\mathrm{DP}}= \frac{N_{\mathrm{GPU}}}{N_{\mathrm{SP}}}.\label{eq:dp_degree}$ The precomputed packing plan assigns a different pack to each replica, while gradients are synchronized across replicas to maintain a consistent global model update. Because the gradient synchronization waits for the slowest replica, the packs are balanced by their estimated FLOPs rather than by sample count or token count (see Section~\ref{sec:train_thr}), reducing workload variation caused by differences in video resolution, duration, and mask density.

\paragraph{FSDP2 hybrid sharding.}
We employ FSDP2~\citep{pytorch_fsdp2} with a logical distributed mesh organized as
\begin{equation}
    [\text{DP replication},\ \text{DP sharding},\
      \text{sequence parallelism}].
\end{equation}
The explicit DP-shard size is set to one. In our implementation, the DP-shard and SP dimensions are combined when constructing the FSDP shard group. Consequently, parameters, gradients, and optimizer states are effectively sharded across the same eight GPUs that form one SP group, and the resulting model shard is replicated across the data-parallel groups.  

This topology avoids performing parameter all-gathers and gradient reduce-scatters over the entire job. These collectives remain within the eight-GPU shard group, while gradient synchronization across replicated groups provides global data-parallel scaling. Parameter all-gathers and gradient reduce-scatters are overlapped with forward and backward computation whenever their dependencies permit. The design therefore trades local communication for substantial reductions in per-GPU model-state memory.

\subsection{GPU Memory Optimization}
\paragraph{Activation checkpointing.}
All main transformer blocks use activation checkpointing~\citep{chen2016training}. Standard checkpointing stores the block input during the forward pass and reconstructs intermediate activations during backpropagation, requiring both the attention and MLP computations to be executed again. Following the principle of selective activation checkpointing~\citep{pytorch2025checkpointing}, we additionally support selective recomputation through \texttt{skip\_attn\_recompute}. When enabled, the attention output and required softmax statistics are retained from the original forward pass. Backward recomputation reuses these saved attention results and recomputes only the remaining block operations, including the MLP. This option preserves attention gradients while avoiding a second execution of the attention forward kernel, trading additional activation memory for lower recomputation cost.

\paragraph{Stage-dependent memory strategy.}
The optimal memory policy depends on the training algorithm. During one-shot training, particularly after the introduction of per-layer camera attention, we enable selective attention checkpointing, asynchronous activation offload, and camera-module checkpointing. Saved attention outputs and checkpoint inputs can be transferred to pinned CPU memory, reducing GPU memory pressure while preserving the compute savings from skipping attention recomputation.

During segment forcing, we instead disable attention-output retention and activation offload. Segment-wise streaming backward already bounds peak memory by releasing intermediate activations as the serial rollout progresses. In this regime, retaining attention outputs increased memory by approximately 1.6\,GB per rank without providing a meaningful speed improvement. We therefore use ordinary attention recomputation together with streaming backward.

\subsection{Training Throughput Optimization}\label{sec:train_thr}
\paragraph{FLOPs-based sequence packing.}
To improve training throughput on long videos with varying spatial resolutions and durations, we employ sequence packing~\citep{krell2022packing} to combine 480p and 720p videos spanning 5--75 seconds into token-budgeted batches. Packing preserves each token's sample identity and original spatial, temporal, and segment metadata, with block-diagonal attention preventing information leaking across videos. Conditioning signals are packed with video tokens to preserve text--segment and camera--frame alignment, while the training loss is computed only over valid video tokens.

Balancing token counts alone does not ensure balanced computation, as segment-wise causal attention incurs substantially different costs depending on the number of unmasked query--key pairs. We therefore estimate the computational cost of each sample directly. Let $n_i$ denote the number of valid video and text tokens of sample $i$, and let
\begin{equation}
    A_i = \sum_{p \in i} \bigl|\{q : M^{(i)}_{pq} \neq -\infty\}\bigr|
    \label{eq:attn_pairs}
\end{equation}
denote the number of unmasked query--key pairs under its segment- and action-aware mask $M^{(i)}$. For segment lengths $m_1,\ldots,m_S$, $A_i$ is dominated by $\sum_{j=1}^{S} m_j \sum_{k \le j} m_k$, which can be computed from segment metadata without materializing the mask. We model the estimated per-sample training FLOPs as
\begin{equation}
    F_i = \alpha n_i + \beta A_i,
    \label{eq:flops_cost}
\end{equation}
where $\alpha$ captures the per-token cost of linear operations across the 54 blocks, including camera cross-attention, and $\beta$ captures the per-pair cost of $QK^\top$ and $PV$ operations. Both coefficients account for backward propagation and activation recomputation, and are derived analytically from the architecture in Table~\ref{tab:architecture_settings} and calibrated against measured step times.

We construct the packing plan offline in two stages. First, samples are grouped into packs satisfying the token budget $\sum_{i \in \mathrm{pack}} n_i \le N_{\max}$, while targeting similar estimated FLOPs per pack. Second, packs are assigned to the $N_{\mathrm{DP}}$ replicas using a longest-processing-time-first greedy algorithm, sorting packs by estimated FLOPs in descending order and assigning each to the least-loaded replica. This balances computational workloads across replicas and reduces synchronization-induced idle time. The resulting plan is stored alongside the data index and loaded by the dataloader without additional runtime scheduling overhead, with consistent packing assignments across ranks within each sequence-parallel group.

\paragraph{Efficient attention with structured masks.} Our training combines segment-causal and action-aware attention with block-diagonal boundaries across packed videos, resulting in complex attention masks over sequences of up to 360K tokens. Dense masks are prohibitively expensive at this scale, while standard FlashAttention kernels~\citep{shah2024flashattention3} cannot efficiently handle such irregular attention patterns. We therefore adopt the Flex-Flash-Attention (FFA) kernel from MagiAttention~\citep{magiattention2025}, which represents structured masks as compact attention slices and skips masked attention blocks. We construct these slices directly from sample, segment, and action metadata without materializing dense masks. We further integrate FFA with eight-way Ulysses sequence parallelism, where each rank processes two of the 16 attention heads over the packed sequence using the same slice specification. This avoids the repeated key--value communication of context-parallel approaches while retaining only two all-to-all collectives per attention layer.

\paragraph{Distributed Muon optimizer.} We train the DiT using Muon~\citep{jordan2024muon,liu2025muon} for the weight matrices in attention and feed-forward layers, and AdamW~\citep{loshchilov2019adamw} for embeddings, biases, and other remaining parameters. To support Muon's full-matrix orthogonalization under FSDP2 parameter sharding, we adopt the distributed Muon implementation from Canzona~\citep{wang2026canzona}.

\paragraph{Asynchronous checkpointing and deterministic resumption.}
We build upon ByteCheckpoint~\citep{wan2024bytecheckpoint} to asynchronously save FSDP2-sharded model and optimizer states, overlapping serialization and storage I/O with training through pinned CPU memory. To ensure deterministic resumption, we additionally checkpoint all dataloader states, including packing-plan positions and sampler states, together with all CPU and GPU RNG states. Restoring these states preserves data ordering and stochastic operations across training interruptions. Furthermore, our global tensor metadata enables checkpoint resharding across different GPU counts and parallel configurations without offline conversion.

\section{Model Training}
\label{sec:hytrain}
We progressively adapt a pretrained bidirectional DiT into an interactive world model through coordinated curricula in causality, interaction, and context length. Section~\ref{sec:hytrain_curriculum} explains the design of this progression, and Sections~\ref{sec:hytrain_cpt}--\ref{sec:hytrain_distillation} describe its implementation, from continual pre-training to few-step autoregressive generation. Section~\ref{sec:hytrain_lora} presents a lightweight recipe for downstream adaptation.

\subsection{Training Curriculum}
\label{sec:hytrain_curriculum}
Our training strategy introduces each capability on top of an already learned foundation. Table~\ref{tab:train_stages} summarizes the five stages, with continual pre-training (CPT) divided into three successive regimes. Together, these stages advance three complementary curricula. 

\paragraph{Causal curriculum.}
We progress from \emph{segment-wise causal attention} to \emph{action-aware causal attention}, followed by \emph{segment forcing} and \emph{diffusion distillation}. Segment-wise masking first establishes the information flow required for autoregressive generation: each segment uses its current instruction and preceding history, while attention remains bidirectional within the segment. Action-aware masking then refines this structure by setting bounded temporal scope for each action instruction, addressing action-order errors and the repetition of earlier actions. Segment forcing subsequently trains the model with  cached context from its own predictions, exposing the model to imperfect histories before inference. Finally, causal distillation reduces the sampling steps while further mitigating error accumulation due to training and inference mismatch. This curriculum progressively convert a bidirectional DiT into an efficient autoregressive model.

\paragraph{Interaction curriculum.}
We extend a \emph{bidirectional, noninteractive text-to-video DiT} into a model supporting \emph{multi-turn free-form text interaction}, and then \emph{free-form text interaction with continuous camera control}. Segment-wise captions and causal attention teach the model to respond to successive text instructions while carrying forward relevant visual context. Camera-conditioned CPT adds geometric control capabilities after the foundation for text-prompt interaction has been established. Mixing text-only and camera-conditioned segments, together with segment-wise camera dropout, exposes the model to transitions between different interaction modes within a sequence. The resulting model can continue an evolving world under either semantic instructions or explicit camera motion.

\paragraph{Context curriculum.}
We combine mixed-resolution and mixed-length training with a progressive increase in the maximum packed sequence length. The context length is set to $180K$ tokens before action-aware CPT and increases to $360K$ tokens from action-aware CPT onward. Mixing resolutions and durations balances spatial detail and temporal coverage; expanding the context length helps the model to understand longer histories and generate richer instruction video sequences. Sequence packing and memory optimizations in Section~\ref{sec:training_infra} make this progression practical. A context length of $360K$ tokens is sufficient to generate minute-scale visual worlds at $720p$ resolution.

\begin{table}[t]
\centering
\small
\setlength{\tabcolsep}{4pt}
\renewcommand{\arraystretch}{1.08}
\vspace{-5mm}
\begin{tabularx}{\linewidth}{@{}c >{\raggedright\arraybackslash}p{0.19\linewidth} >{\raggedright\arraybackslash}X >{\raggedright\arraybackslash}X@{}}
\toprule
Stage & Training regime & Primary data or supervision & Main objective or capability \\
\midrule
1 & Latent-space adaptation & HunyuanVideo~1.5 initialization; 20K steps updating input/output layers, then 60K full-model updates & Transfer HunyuanVideo~1.5 VAE to a more compact DCAE-V latent space. \\
\midrule
2 & Segment-causal CPT & Broad video corpus with segment-wise captions, followed by higher-quality data & Establish segment-wise causal generation before introducing finer-grained action control. \\
\cmidrule(l){2-4}
  & Action-aware CPT & Curated videos with balanced sampling and temporally localized action captions & Learn action timing and cross-segment action isolation; expand the context length from $180K$ to $360K$ tokens. \\
\cmidrule(l){2-4}
  & Camera-conditioned CPT & Videos with and without camera trajectories, mixed at a 1:1 ratio & Add continuous camera control while retaining text-driven generation. \\
\midrule
3 & High-quality annealing & High-quality videos from balanced sources; detailed captions with Kimi K3 and camera trajectory annotations & Refine visual quality, text conditioning, and camera control. \\
\midrule
4 & Segment forcing & A subset of high-quality annealing data & Align training with segment-wise autoregressive inference and its evolving KV-cache state. \\
\midrule
5 & Diffusion distillation & Causal PCM initialization followed by DDMD and adversarial refinement under hybrid forcing & Enable few-step generation, and further reduce errors due to the training-inference mismatch. \\
\bottomrule
\end{tabularx}
\caption{\textbf{Multi-stage training of \shortname.} Stage~2 progresses from segment-causal to action-aware and camera-conditioned CPT. Segment-causal CPT includes a continuation on higher-quality data. Downstream LoRA adaptation is performed separately after the listed training stages.}
\vspace{-3mm}
\label{tab:train_stages}
\end{table}

\subsection{Continual Pre-Training}
\label{sec:hytrain_cpt}

\paragraph{Latent-space adaptation.}
We initialize \shortname from HunyuanVideo~1.5~\citep{wu2025hunyuanvideo} and adapt its VAE to DCAE-V~\citep{chen2025dc}, increasing spatial compression from $16\times16$ to $32\times32$ while keeping temporal compression as a factor of four. This significantly reduces the latent token count at a given video resolution and duration, making training on longer sequences more feasible. We first freeze the intermediate DiT blocks and update only the input projection and output layers for 20K steps, then train the full DiT for another 60K steps.

\paragraph{From segment causality to action alignment.}
CPT first establishes segment-wise causal attention on diverse, variable-length videos with dense structured captions, then continues with higher-quality data under the same attention structure. We subsequently introduce action-aware masking with temporally localized annotations. The key transition is to supervise \emph{when} an instruction applies: descriptive context remains available across segments, whereas action instructions are restricted to their originating segment and annotated temporal windows. Increasing the context length at this stage allows training to include longer action sequences and histories. The attention design is specified in Section~\ref{subsec:causal_masking}.

\paragraph{Camera control.} 
At the end of CPT, we introduce the camera-control interaction capability. Normalized camera extrinsics are encoded as one token per latent frame and injected through frame-wise cross-attention after every MMDiT block. Both the base DiT and the camera modules remain trainable, allowing camera control to be learned jointly with visual quality and prompt adherence. During this training stage, both the camera trajectory and camera-related text descriptions are dropped with $50\%$ probability for each segment. This allows camera-control interaction and text-prompt interaction to coexist within the same video while preventing text-based camera descriptions in captions from leaking the camera motion.

\subsection{High-Quality Annealing}
\label{sec:hytrain_annealing}
After broad capability acquisition, we anneal on high-quality, predominantly single-subject videos from various targeted sources. Kimi K3~\citep{moonshot2026kimik3} is used to generate more accurate dense captions, and we increase the maximum text length from 768 to 1,280 tokens to accommodate more detailed segment descriptions and longer action sequences. The model architecture and flow-matching~\citep{lipman2023flow} objective remain unchanged.

We use different supervision for annealing data according to its camera trajectory quality. Videos whose ViPE trajectories pass validity and motion-variance filters enter the camera-conditioned path; those with unreliable trajectories remain useful as text-conditioned training samples. Camera trajectories are normalized with a shared video-level scale preserving relative motion magnitudes across segments. This concentrates learning on precise text and camera motion supervision while retaining useful videos that lack reliable camera trajectories.

\subsection{Segment Forcing}
\label{sec:hytrain_segment_forcing}
Segment Forcing adapts the model to the evolving history it consumes during autoregressive inference. Inspired by training with model-generated context in Self Forcing~\citep{huang2025selfforcing}, we combine sequential segment-wise denoising, one-step context estimation, and streaming backward. This exposes later segments to prediction errors while retaining direct flow-matching supervision for each current segment.

Given a video divided into multiple segments, the autoregressive DiT processes the segments sequentially. At segment $s$, the model takes the noisy latent $\mathbf{x}_{s,t}$ as input and predicts its flow-matching velocity $\hat{\mathbf{v}}_s$, conditioned on the KV cache constructed from preceding segments and prompts. We apply the flow-matching MSE objective~\citep{lipman2023flow} only to the current segment prediction. The resulting loss is immediately backpropagated, after which the computation graph and intermediate activations of the current segment are released to save GPU memory.

To construct the context for the next segment, we perform one-step denoising from the same noisy latent $\mathbf{x}_{s,t}$ using the predicted velocity:
\begin{equation}
\tilde{\mathbf{x}}_{s,t_{\mathrm{ctx}}}
=
\mathbf{x}_{s,t}
+
(t_{\mathrm{ctx}}-t)\hat{\mathbf{v}}_s,
\qquad
t_{\mathrm{ctx}}\sim\mathcal{U}(0.1,0.3).
\label{eq:seg_forcing_denoise}
\end{equation}
The sampled range approximates the noise levels of context segments encountered during autoregressive inference. The resulting low-noise latent is appended to the KV cache and used as context for subsequent segments. We omit this cache update for the final segment because no later segment consumes it.

\begin{figure}[t]
\centering
\vspace{-7mm}
\includegraphics[width=\linewidth]{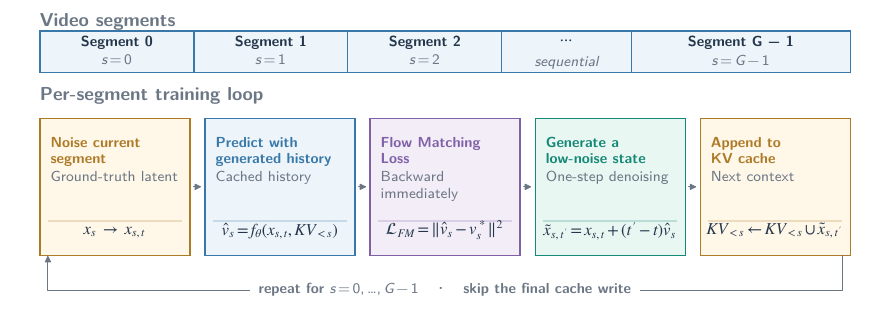}
\vspace{-5mm}
\caption{\textbf{Segment Forcing with streaming backward.} Each current segment receives a flow-matching loss; a one-step low-noise estimation supplies cached context for later segments. Streaming backward retains only one segment’s computation graph at a time, accumulating gradients for a single optimizer step after the full sequence. The cache update of the final segment is skipped.}
\vspace{-3mm}
\label{fig:hytrain_segment_forcing}
\end{figure}

By repeatedly generating and caching its own context, the model is trained under a history distribution that closely matches autoregressive inference. Meanwhile, streaming backward keeps only the latest segment's computation graph in memory. Gradients are accumulated across segments, followed by a single optimizer step after the full sequence has been processed. The complete pipeline is illustrated in Figure~\ref{fig:hytrain_segment_forcing}.

\subsection{Diffusion Distillation}
\label{sec:hytrain_distillation}
\paragraph{Causal distillation.}
Starting from the AR model trained with segment self-forcing, we use two-stage distillation to reduce the sampling budget from 100 to 4 network function evaluations (NFEs) per segment. The two stages establish few-step generation and then refine visual fidelity and motion stability while preserving semantic alignment and camera controllability.

\textit{Stage I: Few-Step Initialization.}
Drawing on consistency-based distillation~\citep{song2023consistency,wang2024phased}, we initialize the student with a causal adaptation of Phased Consistency Models (PCM). The teacher's denoising trajectory is partitioned into phases, and the student learns to map noisy states within each phase to its lower boundary. Training uses teacher forcing: each segment is conditioned on ground-truth history, text, and camera inputs. The resulting few-step checkpoint initializes Stage II.

\textit{Stage II: Quality Refinement.}
Starting from the PCM checkpoint, we refine the student under hybrid forcing~\citep{hong2025relic} to improve visual fidelity and motion stability while preserving semantic alignment. For longer sequences, a noise-perturbed ground-truth prefix initializes the student's causal KV cache. The student then generates the suffix autoregressively with 4 NFEs per segment, using each generated segment as context for the next. This combines ground-truth context with training on the student's own generated history. Shorter sequences are generated entirely by the student.

We jointly optimize the student with a causal adaptation of Decoupled Distribution Matching Distillation (DDMD) and an adversarial hinge objective, drawing on distribution-matching distillation~\citep{liu2025decoupled,yin2024dmd,yin2024dmd2} and adversarial distillation~\citep{sauer2023adversarial,lin2025apt,lin2025aapt2,cheng2026vpae,cheng2026rewardlightning}, respectively. DDMD samples noise levels independently for CFG and DM, restricting CFG to the remaining denoising interval while sampling DM over the full interval~\citep{liu2025decoupled}. DDMD supervision is applied only to the final generated segment, with gradients stopped through earlier segments and their cached representations. The teacher remains frozen, while the fake-score model is trained with flow matching~\citep{lipman2023flow} on detached student samples. The adversarial objective complements DDMD with supervision from high-quality real videos to improve visual fidelity and motion stability.

\paragraph{Causal inference.}
At inference, the distilled student generates segments autoregressively with 4 NFEs per segment, conditioned on text, camera inputs, and previously generated history. After each segment is generated, we perturb it with flow noise at $\sigma=0.206897$ before updating the causal KV cache for subsequent segments.

\subsection{Downstream Adaptation via LoRA}
\label{sec:hytrain_lora}
\captionsetup[figure]{font=small}

\definecolor{Navy}{HTML}{24364B}
\definecolor{Purple}{HTML}{8061A7}
\definecolor{PurpleFill}{HTML}{F2EDF7}
\definecolor{Blue}{HTML}{3978A8}
\definecolor{BlueFill}{HTML}{EDF5FA}
\definecolor{Teal}{HTML}{168876}
\definecolor{TealFill}{HTML}{E9F7F3}
\definecolor{GrayFill}{HTML}{EEF2F6}

\begin{figure}[t]
  \centering
  \resizebox{\linewidth}{!}{%
    \begin{tikzpicture}[
      font=\sffamily\small,
      text=Navy,
      box/.style={
        draw=Navy,
        fill=GrayFill,
        rounded corners=4pt,
        minimum width=3.6cm,
        minimum height=1.05cm,
        align=center,
        line width=0.9pt,
        font=\sffamily\small
      },
      arrow/.style={
        -{Stealth[length=5pt]},
        draw=Navy,
        line width=0.9pt
      }
    ]
      \node[box] (base) at (0,1.1)
        {Non-distilled\\checkpoint};

      \node[
        box,
        draw=Purple,
        fill=PurpleFill,
        text=Purple
      ] (lora) at (6.9,1.1)
        {LoRA weights};

      \node[
        box,
        draw=Blue,
        fill=BlueFill,
        text=Blue
      ] (distilled) at (6.9,-1.1)
        {Distilled\\checkpoint};

      \node[
        circle,
        draw=Navy,
        fill=GrayFill,
        minimum size=1cm,
        line width=0.9pt,
        font=\sffamily\LARGE
      ] (merge) at (10,0) {+};

      \node[
        box,
        draw=Teal,
        fill=TealFill,
        text=Teal
      ] (output) at (13.6,0)
        {Merged checkpoint\\[-1pt]
         {\footnotesize for inference}};

      \draw[arrow,draw=Purple]
        (base.east) --
        node[
          midway,
          above=5pt,
          align=center,
          font=\sffamily\small,
          text=Purple
        ] {Segment forcing\\fine-tuning}
        (lora.west);

      \draw[arrow,draw=Purple,rounded corners=8pt]
        (lora.east) --
        node[
          midway,
          above=5pt,
          font=\sffamily\small,
          text=Purple
        ] {Merge}
        (merge.north |- lora.east) --
        (merge.north);

      \draw[arrow,draw=Blue,rounded corners=8pt]
        (distilled.east) -| (merge.south);

      \draw[arrow,draw=Teal]
        (merge.east) -- (output.west);
    \end{tikzpicture}%
  }
  \caption{\textbf{Downstream LoRA adaptation workflow.} Train LoRA adapters
    on the base autoregressive model (before distillation), then merge the learned weights into
    the distilled model for efficient deployment, without
    full-model fine-tuning or repeating distillation.}
  \label{fig:lora_workflow}
\end{figure}
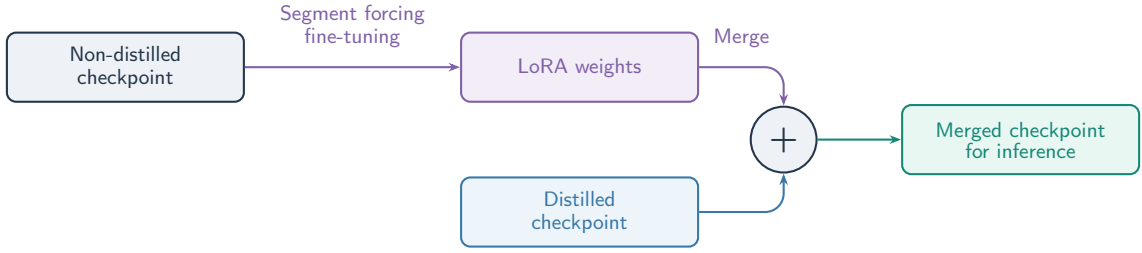
To help users customize our model for their use cases, we provide a recommended workflow based on Low-Rank Adaptation (LoRA)~\citep{hu2021lora}. We present this workflow as our best-practice recipe for downstream adaptation and encourage users to follow it when developing their own specialized models.

\paragraph{Best practices for downstream adaptation.} We recommend training LoRA adapters on the base autoregressive model before distillation, then merging the learned LoRA adapter weights into the distilled model for deployment. The workflow is illustrated in Figure~\ref{fig:lora_workflow}. Users can follow this recipe with on the order of 100 task-specific samples selected to reflect their intended visual style, usage scenario, or application domain. This workflow enables lightweight customization without full-model fine-tuning or repeating the distillation pipeline, while retaining the inference efficiency of the distilled model. The following examples illustrate these best practices and provide practical references for users adapting the model to their own applications.

\paragraph{Cinematic style LoRA.}
As an example of visual-style adaptation, we train a cinematic-style LoRA adapter on the base autoregressive model using cinematic videos paired with high-quality captions from K3 as our SFT data. We set the LoRA rank to 256 and alpha to 512, and train for 500 iterations with a learning rate of $10^{-4}$. For effective domain adaptation, we recommend using a slightly higher learning rate than that used for base-model training. Due to the small size of the SFT dataset, we employ sequence packing in the data loader such that each training iteration covers the entire dataset, making 500 iterations equivalent to 500 epochs. We then merge the resulting LoRA adapter weights into the distilled model for inference. 

\paragraph{Robotics LoRA.}
Beyond visual-style adaptation, robotics is another potential application of this recipe. We perform task-specific robotics LoRA fine-tuning on a set of robotics dual-arm egocentric task videos~\citep{realsourceworld}, which is not overlapped with our training set. We conduct the LoRA tuning for approximately $1,500$ epochs on the selected 128-sample tuning set with a LoRA rank of 512.

In Section~\ref{sec:lora_results}, we provide a preliminary evaluation of the LoRA-adapted model against the distilled model before merging the LoRA weights.

\section{Model Inference}
Interactive deployment requires translating user intent into suitable conditioning inputs and generating responses efficiently. To address these complementary requirements, we develop a VLM-based Prompt Enhancer (PE) and optimize the model's inference pipeline.

The PE translates user intent into structured instructions for the video generator. Given previously generated video segments, the current user instruction, and optional camera-control signals, the PE produces a prompt for the next segment. This design builds on recent approaches to streaming text control and agent-mediated world interaction~\citep{hyintro_longlive_2025,gao2026infinite}, while using a structured caption format aligned with our training annotations.

\subsection{Prompt Enhancer}
\label{sec:prompt_engineering}

\begin{figure*}[t]
\centering
\vspace{-5mm}
\resizebox{0.95\linewidth}{!}{%
\begingroup%
\DeclareFontFamily{T1}{pess}{}%
\DeclareFontShape{T1}{pess}{m}{n}{<-> ec-lmss8}{}%
\DeclareFontShape{T1}{pess}{bx}{n}{<-> ec-lmssbx10}{}%
\definecolor{peInk}{HTML}{24364B}%
\definecolor{peNote}{HTML}{6C7785}%
\definecolor{peLine}{HTML}{C7D0D9}%
\definecolor{peFrame}{HTML}{F6F8FA}%
\definecolor{peCore}{HTML}{EEF2F6}%
\definecolor{pePurple}{HTML}{8061A7}%
\definecolor{pePurpleFill}{HTML}{F2EDF7}%
\definecolor{peBlue}{HTML}{3978A8}%
\definecolor{peBlueFill}{HTML}{EDF5FA}%
\definecolor{peTeal}{HTML}{168876}%
\definecolor{peTealFill}{HTML}{E9F7F3}%
\definecolor{peAmber}{HTML}{AD7A28}%
\definecolor{peAmberFill}{HTML}{FFF7E8}%
\definecolor{peSky}{HTML}{CFE3F3}%
\definecolor{peMountain}{HTML}{B9C8C2}%
\definecolor{peHill}{HTML}{93AE9A}%
\definecolor{peTrail}{HTML}{E6D6BC}%
\definecolor{peTree}{HTML}{5E8569}%
\newcommand{\PEText}[7]{%
  \pgfmathsetmacro{\PELeading}{#3*1.22}%
  \node[inner sep=0,outer sep=0,anchor=#6,align=center,text=#4,%
    font={\fontencoding{T1}\fontfamily{pess}\fontseries{#5}%
      \fontsize{#3}{\PELeading}\selectfont}]%
    at (#1,#2) {#7};}%
\newcommand{\PECard}[6]{%
  \path[fill=#6,draw=#5,line width=.85bp,rounded corners=4bp]%
    (#1,#2) rectangle (#3,#4);}%
\newcommand{\PEVideo}[3]{%
  \begin{scope}[shift={(#1,#2)},scale=#3]%
    \foreach \k in {0,1,2}{%
      \begin{scope}[shift={({-62+4*\k},{-32-4*\k})}]%
        \path[fill=white,draw=peBlue!32,line width=.6bp,rounded corners=2bp]%
          (-3,-3) rectangle (119,73);%
        \begin{scope}%
          \clip (0,0) rectangle (116,70);%
          \fill[peSky!65] (0,0) rectangle (116,70);%
          \fill[peMountain] (0,42)--(21,12)--(43,33)--(69,14)--(99,36)%
            --(116,26)--(116,70)--(0,70)--cycle;%
          \fill[peHill] (0,47)--(27,31)--(49,45)--(78,32)--(116,47)%
            --(116,70)--(0,70)--cycle;%
          \fill[peTrail] (44,70)--(59,38)--(65,38)--(84,70)--cycle;%
          \foreach \tx/\ty/\th in {8/50/16,18/44/14,31/54/12,83/44/12,95/49/16,106/55/17}{%
            \fill[peTree] (\tx,{\ty-\th})--({\tx-3.8},\ty)--({\tx+3.8},\ty)--cycle;%
            \draw[peTree,line width=1bp] (\tx,\ty)--(\tx,{\ty+3});%
          }%
        \end{scope}%
      \end{scope}%
    }%
  \end{scope}}%
\newcommand{\PECamera}{%
  \begin{scope}[shift={(32,321)}]%
    \foreach \gx in {0,34,68,102,136,164}{%
      \draw[peLine!45,line width=.55bp] (80,-1)--(\gx,34);}%
    \foreach \gy in {15,24,34}{%
      \draw[peLine!45,line width=.55bp] ({80-2.4*\gy},\gy)--({80+2.4*\gy},\gy);}%
    \draw[flow,peTeal,dash pattern=on 4bp off 3bp]%
      (10,30)..controls(60,22)and(100,11)..(147,1);%
    \foreach \cx/\cy in {10/30,51/22,92/13,129/5}{%
      \path[fill=peTealFill,draw=peTeal,line width=.65bp]%
        (\cx,\cy)--({\cx+7},{\cy-5})--({\cx+8},{\cy+1})--cycle;}%
  \end{scope}}%
\begin{tikzpicture}[x=1bp,y=-1bp,line cap=round,line join=round,
  flow/.style={draw=peNote,line width=1bp,rounded corners=3bp,
    -{Triangle[length=4.3bp,width=3.8bp]}},
  feedback/.style={flow,draw=peNote!80,dash pattern=on 5bp off 3.5bp}]
\path[use as bounding box] (0,0) rectangle (1048,510);
\fill[white] (0,0) rectangle (1048,510);

\foreach \top/\bottom in {10/174,194/380}{
  \path[fill=peFrame,draw=peLine,line width=.8bp,rounded corners=7bp,
    dash pattern=on 5bp off 4bp] (8,\top) rectangle (1016,\bottom);}
\PEText{24}{32}{17}{peInk}{bx}{west}{First turn}
\PEText{24}{213}{17}{peInk}{bx}{west}{Subsequent turn}

\PECard{24}{67}{205}{139}{peLine}{white}
\PEText{114.5}{89}{15}{peInk}{m}{center}{Subject: ...}
\PEText{114.5}{118}{15}{peInk}{m}{center}{Environment: ...}
\PECard{261}{65}{361}{141}{pePurple}{pePurpleFill}
\PEText{311}{103}{18}{pePurple}{bx}{center}{Prompt\\Enhancer}
\PECard{415}{58}{685}{148}{peLine}{white}
\PEText{550}{78}{15.5}{peInk}{bx}{center}{Video-level info}
\draw[peLine!60,line width=.6bp] (429,94)--(671,94);
\PEText{550}{109}{15}{peInk}{m}{center}{Description long}
\PEText{550}{131}{15}{peInk}{m}{center}{Contained actions}
\PECard{744}{68}{844}{138}{peAmber}{peAmberFill}
\PEText{794}{103}{19}{peAmber}{bx}{center}{AR-DiT}
\PEText{942}{52}{15.5}{peInk}{bx}{center}{Video segment 1}
\PEVideo{942}{107}{.88}
\draw[flow] (205,103)--(261,103);
\draw[flow] (361,103)--(415,103);
\draw[flow] (685,103)--(744,103);
\draw[flow] (844,103)--(882,103);

\PEText{24}{235}{15.5}{peBlue}{bx}{west}{Text Prompt}
\PECard{24}{248}{205}{289}{peBlue}{peBlueFill}
\PEText{114.5}{268}{15}{peBlue}{m}{center}{Prompt: ...}
\PEText{24}{306}{15.5}{peTeal}{bx}{west}{Camera Control}
\PECard{24}{319}{205}{364}{peTeal}{peTealFill}
\PECamera
\PECard{261}{259}{361}{348}{pePurple}{pePurpleFill}
\PEText{311}{303.5}{18}{pePurple}{bx}{center}{Prompt\\Enhancer}
\PECard{415}{244}{685}{294}{peBlue}{peBlueFill}
\PEText{550}{258}{15}{peInk}{m}{center}{Description change}
\PEText{550}{280}{15}{peInk}{m}{center}{Contained actions}
\PECard{415}{310}{685}{364}{peTeal}{peTealFill}
\PEText{550}{325}{15}{peInk}{m}{center}{Description change}
\PEText{550}{349}{15}{peInk}{m}{center}{Camera Trajectory}
\PECard{744}{264}{844}{347}{peAmber}{peAmberFill}
\PEText{794}{305.5}{19}{peAmber}{bx}{center}{AR-DiT}
\PEText{942}{253}{15.5}{peInk}{bx}{center}{Video segment t}
\PEVideo{942}{307}{.88}
\draw[flow,peBlue] (205,268)--(261,268);
\draw[flow,peTeal] (205,338)--(261,338);
\draw[flow,peBlue] (361,268)--(415,268);
\draw[flow,peTeal] (361,338)--(415,338);
\draw[flow,peBlue] (685,268)--(744,268);
\draw[flow,peTeal] (685,338)--(744,338);
\draw[flow,peBlue] (844,283)--(882,283);
\draw[flow,peTeal] (844,329)--(882,329);

\PEText{24}{400}{15.5}{peInk}{bx}{west}{Full interaction history}
\PECard{24}{414}{214}{500}{peLine}{white}
\PEText{119}{428}{12.5}{peNote}{m}{center}{All prior rounds: 1 ... t - 1}
\path[fill=pePurpleFill,draw=pePurple!65,line width=.65bp,rounded corners=2bp]
  (42,442) rectangle (82,463);
\path[fill=white,draw=pePurple,line width=.65bp,rounded corners=2bp]
  (49,449) rectangle (91,473);
\path[fill=white,draw=pePurple,line width=.65bp]
  (55,473)--(55,478)--(63,473);
\foreach \yy in {455,461,467}{
  \draw[pePurple!65,line width=.6bp] (55,\yy)--(83,\yy);}
\PEVideo{166}{459}{.44}
\PEText{68}{489}{12.5}{peNote}{m}{center}{Dialogue}
\PEText{166}{489}{12.5}{peNote}{m}{center}{Videos}
\draw[flow] (214,449)--(311,449)--(311,348);

\draw[draw=peNote!80,line width=1bp,rounded corners=3bp,
  dash pattern=on 5bp off 3.5bp] (997,103)--(1032,103)--(1032,471);
\draw[draw=peNote!80,line width=1bp,dash pattern=on 5bp off 3.5bp]
  (997,303)--(1032,303);
\fill[peNote!80] (1032,303) circle[radius=1.8bp];
\PECard{462}{450}{838}{492}{peLine}{peCore}
\PEText{650}{471}{15}{peInk}{m}{center}{Append this round's dialogue + video}
\draw[feedback] (1032,471)--(838,471);
\draw[feedback] (462,481)--(214,481);
\end{tikzpicture}%
\endgroup
}
\caption{\textbf{Prompt Enhancer (PE) workflow}. PE generates an initial dense caption in the first turn. In subsequent turns, users can choose text prompts or camera control, and PE combines the selected input with the full interaction history to generate a dense caption that AR-DiT uses to produce the next video segment. Each turn’s dialogue and generated video are appended to the history.}
\label{fig:pe_workflow_v2}
\end{figure*}
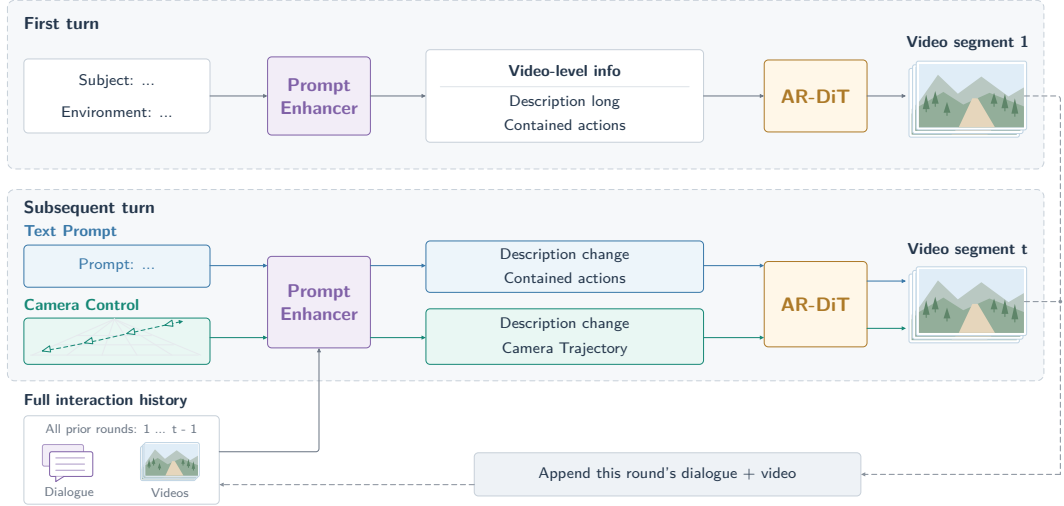
\subsubsection{Prompt Enhancer Workflow}

\paragraph{Multi-turn interaction.}
Prompt Enhancer (PE) converts user instructions into structured dense captions for
segment-by-segment video generation. At each turn, PE produces one caption
for the next segment, which the DiT then uses to generate the corresponding
video. The interaction proceeds causally: each new turn conditions on the
current user input and the complete history accumulated so far.

Formally, the caption at turn $t$ is generated as
\begin{equation}
    y_t \sim p_{\theta}\!\left(
        \cdot \mid S, u_t, H_{t-1}, V_{<t}, m_t
    \right),
    \label{eq:incremental_prompt_rewriting}
\end{equation}
where $S$ is the system prompt, $u_t$ is the current user input, and $m_t$
specifies the rewriting mode. The dialogue history $H_{t-1}$ contains
all preceding user requests and PE responses, while
$V_{<t} = (V_1,\ldots,V_{t-1})$ contains all previously generated video
segments. Both histories are empty at the first turn. After each round,
the dialogue and generated video are retained as context for subsequent
turns.

\paragraph{Initial and incremental captions.}
The output format depends on the turn and the selected control mode:
\begin{itemize}
[leftmargin=10pt]
    \item \textbf{First-turn initialization.}
    The user provides subject and environment descriptions, together with
    any specified visual style and viewing perspective. PE returns
    video-level metadata and the first segment caption, including
    \texttt{visual\_description\_long} to establish the scene and
    \texttt{contained\_actions} to specify subject actions, according to the caption format defined in Figure~\ref{fig:caption_schema}.

    \item \textbf{Subsequent turns in text prompt mode.}
    PE combines the current user description with the full dialogue and
    video history to produce an incremental dense caption. The output
    contains \texttt{description\_change} for scene and camera updates
    relative to the established context, and \texttt{contained\_actions}
    for the next segment's subject behavior.

    \item \textbf{Subsequent turns in camera control mode.}
    PE outputs only \texttt{description\_change}, leaving
    \texttt{contained\_actions} empty. The user-defined camera trajectory
    is supplied separately to DiT alongside this incremental caption.
\end{itemize}

\subsubsection{Prompt Enhancer Training}
To support local deployment, we train a lightweight PE model using a dataset of multi-turn multimodal conversations. Each turn pairs the current user instruction and previous dialogue and video history with a structured dense caption response. We construct the data in two ways. First, we synthesize user prompts from existing dense captions, checking that the prompts preserve the specified subject attributes, scene content, and action ordering. Second, we collect real user prompts as seed data and augment them into more synthetic user prompts. We then run these user prompts through the complete PE and video generation pipeline, recording the caption response and generated video at each turn. Each PE turn uses the current instruction and the full dialogue and video history from previous turns. The dataset includes scene initialization and follow-up interactions across different subjects, environments, viewpoints, and control modes. We also add examples to expand the coverage of action types, different action combinations, and temporal orderings, and check for inconsistencies between instructions, captions, and successive segments. 
We use Qwen3.8-27B as our base model and train our PE through supervised fine-tuning. 

\subsection{Inference Optimization}
To make our model accessible to users with limited GPU resources, we optimize both autoregres-
sive generation and video decoding for deployment on a single NVIDIA RTX 4090 GPU. We com-
bine INT8 storage, sliding-window attention (SWA), and CPU offloading to reduce GPU memory
usage, and use sparse attention to improve generation efficiency. For video decoding, we train
a PixelDiT decoder based on PixelUMM~\citep{wei2026pixelumm} and optimize it for single-GPU inference.

\paragraph{Efficient single-GPU inference on NVIDIA RTX 4090.}
We reduce inference memory primarily through INT8 quantization, sliding-window attention (SWA), and CPU offloading. We quantize model weights and cache historical attention keys and values in INT8 with floating-point scales, dequantizing them to BF16 during computation while keeping the current segment's attention inputs in BF16. SWA maintains a bounded history on the GPU, preventing the KV cache from growing with rollout duration. We further offload inactive model components to CPU memory between pipeline stages, such as offloading the VAE during DiT denoising and loading the text encoder only when needed. This reduces peak memory usage at the cost of CPU--GPU transfer latency. Peak GPU memory also depends on the current segment length and temporary computation buffers. For example, generating a 100-second video across seven segments at 1280$\times$704 resolution and 24 FPS requires 20.53 GiB of peak GPU memory. To further accelerate inference, we use dense attention for the first two denoising steps, followed by sparse attention that retains 25\% of candidate video blocks in subsequent steps. These optimizations may introduce subtle differences that become more noticeable over successive segments during autoregressive generation.

\paragraph{PixelUMM decoder adaptation.}
We replace the DCAE-V decoder with a pixel-space diffusion decoder initialized from the vision part of PixelUMM~\citep{wei2026pixelumm}. The decoder takes the DCAE-V latents as a condition and recovers the corresponding video from pixel-space Gaussian noise. When reconstructing videos with complex natural textures, the original DCAE-V decoder, which operates on highly compressed latents, tends to output a repetitive grid-like pattern in place of the true texture. In contrast, our decoder uses its pretrained generative prior to restore realistic high-frequency details. The decoder is trained with PixelUMM's diffusion loss, first on images and 40-frame videos at $256\times256$, and then on images and longer videos at 384p--720p. For long videos, we adopt the temporal tiling strategy of the H3 decoder~\citep{minimax2026h3} rather than that of the DCAE-V decoder, decoding overlapping chunks of 40 frames with a 20-frame overlap blended by cosine weights. Spatially, instead of overlapping tiling, which incurs substantial redundant computation, we decode each frame in a single pass and use sliding-tile attention~\citep{zhang2025fast}, in which each patch attends only to the patches in a local spatial neighborhood across all frames in the chunk. Finally, we distill the decoder with DMD2~\citep{yin2024dmd2} for one-step inference, and accelerate its inference on a single NVIDIA RTX 4090 GPU with FP8 quantization.

\section{Evaluation}
\label{sec:evaluation}
\begin{figure*}[t]
    \centering
    \includegraphics[width=\textwidth]{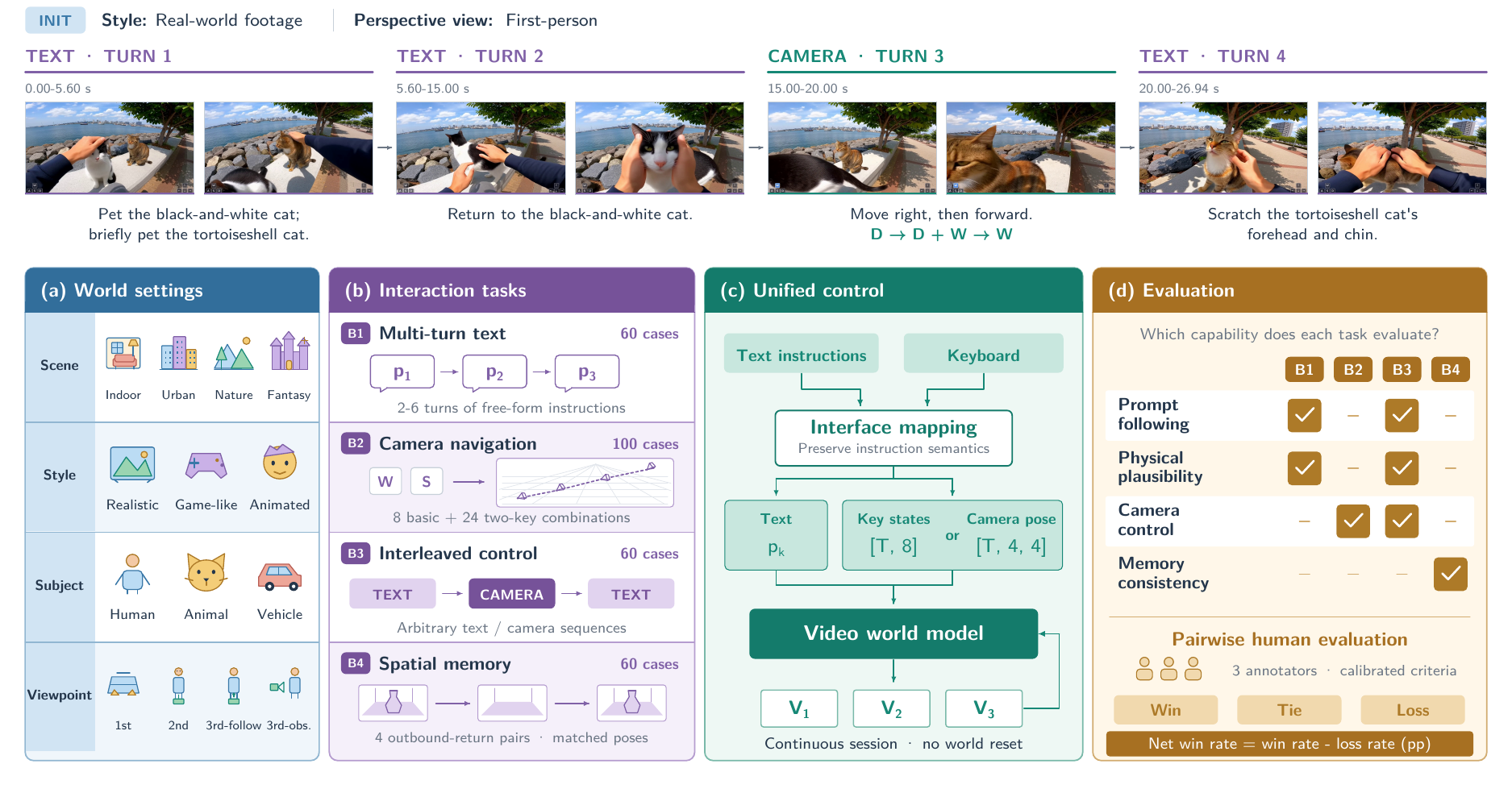}
    \caption{\textbf{Overview of Hy-Nomas-Bench.}
    \textbf{Top:} Model outputs for an example case from Bench-3 (interleaved text--camera control).
    \textbf{Bottom:} (a) World settings span diverse scenes, visual styles,
    subjects, and viewpoints. (b) 280 cases cover multi-turn text interaction, camera navigation,
interleaved control, and spatial memory.
(c) Unified control maps instructions to text, binary key states
$[T,8]$ (\texttt{W/A/S/D} and four arrow keys), or per-frame
$4\times4$ camera-to-world matrices $[T,4,4]$, where $T$ is the
frame count. Generation continues across turns without world resets.
(d) Three annotators evaluate prompt following, physical plausibility,
camera control, and memory consistency through pairwise comparisons.
Results are summarized by win/tie/loss rates and net win rate
(win minus loss, in percentage points).}
    \label{fig:hy-nomas-bench-overview}
\end{figure*}

To systematically evaluate interaction capability and memory
consistency of world models, we introduce \textbf{Hy-Nomas-Bench}, comprising
280 test cases across four benchmarks: multi-turn text interaction,
camera navigation, interleaved text--camera control, and spatial memory. We compare \shortname{} with three representative interactive world models: Genie~3~\citep{deepmind2025genie3}, LingBot-World~v2~\citep{gao2026infinite}, and Happy Oyster~\citep{happyoyster2026}. We generate outputs using each model's supported interaction interface and conduct pairwise human evaluation on prompt following, physical plausibility, camera control, and memory consistency.

\subsection{Hy-Nomas-Bench and Metrics}
The overview of Hy-Nomas-Bench is shown in
Figure~\ref{fig:hy-nomas-bench-overview}. We construct four benchmarks totaling 280 test cases
that assess model capabilities through free-form text prompts, keyboard-based camera commands, and interleaved text and camera inputs, as well as spatial memory across successive interactions. The benchmarks cover first-person, second-person, third-person following, and third-person observer viewpoints in photorealistic, game-like, and animated environments. We release original user prompts for evaluating other models through their native interfaces, alongside PE-enhanced prompts following our structured caption format to reproduce our results \footnote{\url{https://huggingface.co/datasets/tencent/Hy-Nomas-Bench}}. Camera controls provide frame-aligned keyboard states and reference poses with explicit timing and motion conventions. All four benchmarks cover complex multi-turn interactions with cases that require the model to generate videos longer than $60$ seconds.

We consider four dimensions to evaluate world models. \textit{Prompt following} measures adherence to free-form text instructions; \textit{physical plausibility} assesses the realism and consistency of generated physical interactions; \textit{camera control} measures how accurately the model responds to navigation commands; and \textit{memory consistency} assesses spatial coherence when revisiting previously observed environments. These criteria are designed to cover different capabilities of world models, \ie, instruction execution, plausible dynamics, navigation, and spatial recall. Conventional evaluation metrics used in video generation are not considered in Hy-Nomas-Bench (such as visual fidelity, aesthetics, audio quality, \etc), as we focus on capabilities that are crucial to world models.

A closely related work is WBench~\citep{ying2026wbench}, which evaluates text-prompt and camera-control capability separately, since most existing models can only handle one type of interaction.  In contrast, we introduce interleaved text and camera control within a single continuous test case. We also assess spatial memory through specially designed camera trajectories that leave and return to an earlier viewpoint, testing whether previously observed anchor objects and environments remain present and consistent. We describe each benchmark below.

\paragraph{Bench-1: Multi-turn free-form text.}
This benchmark examines whether instruction execution remains consistent with an evolving scene across successive turns. We construct 60 scenarios with 2--6 turns, using prompts curated by human annotators. The sequences include interactions that depend on entities and states established in preceding turns, requiring models to execute new instructions while retaining relevant scene context. This introduces dependencies across turns beyond isolated action completion: previously introduced subjects must remain identifiable, object interactions must remain compatible with the current scene, and action transitions must remain physically plausible. The resulting sequences probe both local instruction adherence and the persistence of scene content throughout the whole video.

\paragraph{Bench-2: Camera navigation.}
This benchmark targets the combination of different keyboard actions and reversibility of camera controls. The 100 cases cover all eight basic keyboard actions and all 24 two-key combinations that exclude opposing inputs along the same control axis. These combinations encompass simultaneous translations, simultaneous rotations, and coupled translation--rotation motions, testing whether individually meaningful controls remain correctly composed when applied together. Each segment holds a fixed action for 2--4 seconds and is followed by an equal-duration inverse action. Under the specified motion-integration rules, the reference trajectories return to the same position and orientation at each pair boundary. This construction provides controlled motion intervals and known return poses for examining directional adherence, coupled-motion errors, and drift over successive outbound--return pairs.

\paragraph{Bench-3: Interleaved text and camera control.}
This benchmark tests whether control responsiveness and scene continuity persist when switching between text prompts and camera control within a rollout. The 60 cases contain 2--7 segments, each specifying either a free-form semantic interaction or an explicit camera-control sequence. Bench-3 covers both text-to-camera and camera-to-text transitions. Text segments introduce actions and scene changes, while camera segments navigate the scene under prescribed camera trajectories. It requires models to preserve the consequences of earlier interactions during navigation and interpret subsequent text instructions correctly to update the scene. World models need to respond correctly when switching between different control signals.

\begin{figure*}[t]
    \centering
    \includegraphics[width=0.9\textwidth]{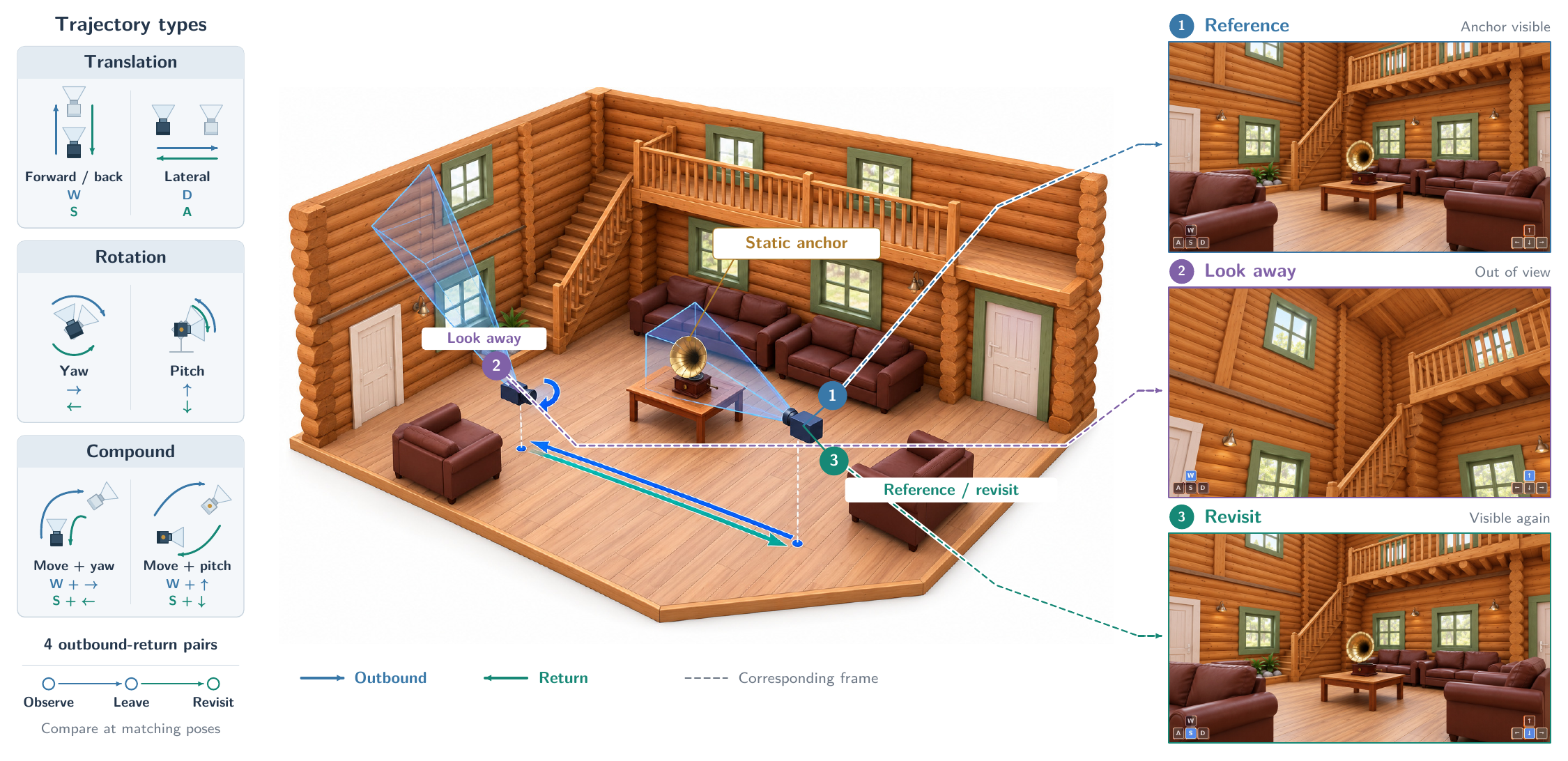}
    \caption{\textbf{Illustration of the design of Bench-4: Spatial memory.}
Each uses a static
anchor object and four outbound--return pairs. Bench-4
evaluates object permanence under different camera movements, and appearance/layout consistency.
\textbf{Left:} Different trajectory types.
\textbf{Center:} Scene, camera frustum, and trajectory.
\textbf{Right:} Inference results by \shortname{}.}
    \label{fig:bench4_spatial_memory}
\end{figure*}

\paragraph{Bench-4: Spatial memory.}
This benchmark evaluates whether previously observed content remains consistent after leaving the field of view. We construct 60 first-person scenarios, evenly distributed across real-world and game-style indoor and outdoor environments. Each scenario contains a distinctive static anchor object and eight camera-control segments forming four outbound--return pairs, covering translations, rotations, and more complex camera motions. A static anchor object is used as a reference to identify any memory inconsistency. The camera trajectories are designed to take the anchor out of view before returning to a previously observed viewpoint. The task of Bench-4 is to compare initial and revisited frames at matched camera positions and orientations, and assess the differences between the frames. We focus on the anchor's identity and appearance, as well as the surrounding scene layout when measuring memory consistency. Figure~\ref{fig:bench4_spatial_memory} illustrates the anchor-based camera trajectory designs.

\subsection{Evaluation Protocol}
\paragraph{Evaluation setup.}
To evaluate different models with heterogeneous interaction interfaces, we adapt Hy-Nomas-Bench to each model’s native interaction setting while preserving the designed task within each benchmark. Models that require an initial image are initialized with the first frame generated by Hy-Nomas, ensuring a consistent visual starting point. Unless stated otherwise, all models are evaluated at their original resolutions using their default inference settings.

\shortname{} supports interleaved text and camera control within one unified model. For LingBot-World~v2~\citep{gao2026infinite}, we adapt their open-source model for free-form text and interleaved evaluation. Genie~3~\citep{deepmind2025genie3} is only evaluated on camera navigation and spatial memory, as the official API does not support text-prompt based interactions. Happy Oyster~\citep{happyoyster2026} is evaluated separately in \emph{Directing} mode for text interaction and \emph{Wandering} mode for navigation, and it does not support interleaved text and camera control within the same video. Table~\ref{tab:baseline-interfaces} summarizes the evaluation settings across different models. Due to the interaction limitations of each model, some benchmarks are not applicable to certain models.

\begin{table*}[t]
    \centering
    \small
    \setlength{\tabcolsep}{6pt}
    \begin{tabular}{@{}lcccc@{}}
        \toprule
        System
        & \shortstack{Multi-turn\\free-form text}
        & \shortstack{Camera\\navigation}
        & \shortstack{Interleaved text\\and camera}
        & \shortstack{Spatial\\memory} \\
        \midrule
        \shortname{}
        & $\checkmark$ & $\checkmark$ & $\checkmark$ & $\checkmark$ \\
        Genie~3
        & -- & $\checkmark$ & -- & $\checkmark$ \\
        LingBot-World~v2
        & $\checkmark$ & $\checkmark$ & $\checkmark$ & $\checkmark$ \\
        Happy Oyster
        & $\checkmark$ & $\checkmark$ & -- & $\checkmark$ \\
        \bottomrule
    \end{tabular}
    \caption{
        \textbf{Models evaluated on each benchmark.} Checkmarks indicate the model is evaluated on the benchmark; dashes
        indicate that the benchmark settings is not applicable to the model's provided interaction interface.
        For LingBot-World~v2,  we adapt their open-source model for Bench-3.
    }
    \vspace{-3mm}
    \label{tab:baseline-interfaces}
\end{table*}

\begin{figure}[t]
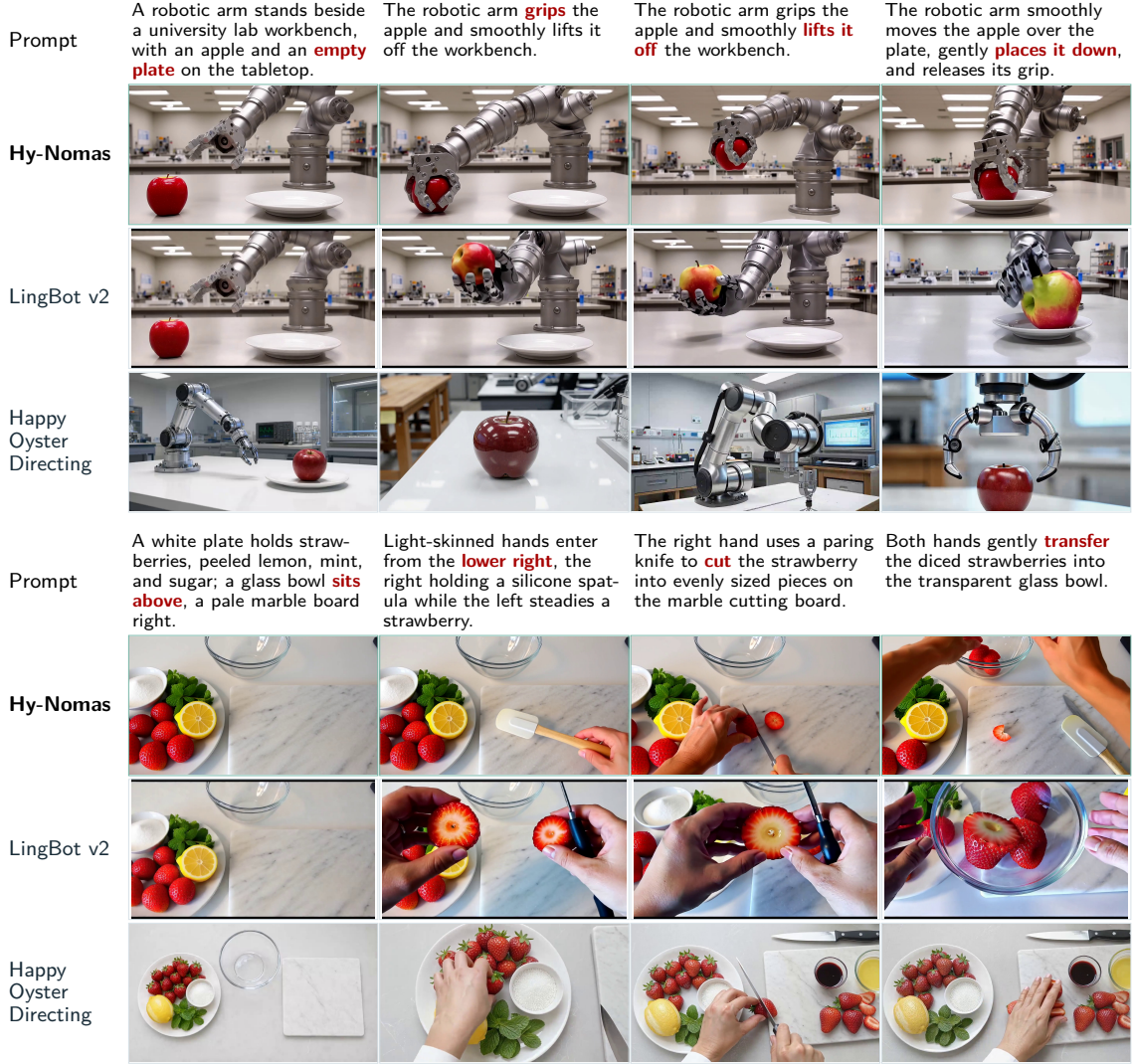

    \centering
    \begingroup
    \newcommand{\QualitativeFont}{\fontfamily{lmss}\selectfont}
    \resizebox{\linewidth}{!}{\input{figures/qualitative/text_robot/body}} 
    
    \vspace{0.1in}
    \resizebox{\linewidth}{!}{\input{figures/qualitative/text_airplane/body}}
    \endgroup
    \caption{\textbf{Qualitative comparison of multi-turn text interaction.}
In the first example, \shortname{} clearly shows
the apple being grasped, lifted, and brought to the plate.
LingBot-World~v2 changes the apple's color and gripper structure, while
HappyOyster-Directing shows discontinuous views without a clear
grasp-and-lift progression.
In the second example, \shortname{} visualizes the requested spatula, cutting on
the board, and strawberries in the glass bowl. LingBot-World~v2 only shows
cutting strawberries and a knife without spatula, while
HappyOyster-Directing does not move strawberries into the bowl.
}
\label{fig:qual_text_robot}
\end{figure}
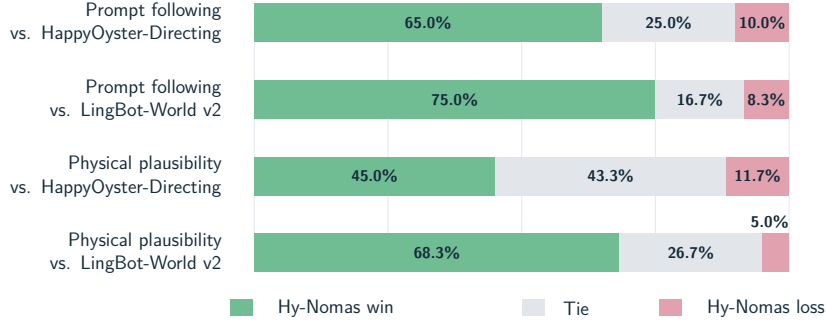
\begin{figure*}[t]
  \centering
  \definecolor{figwin}{HTML}{70BC93}
  \definecolor{figtie}{HTML}{E2E6EB}
  \definecolor{figloss}{HTML}{E2A1AF}
  \definecolor{figink}{HTML}{152A3A}
  \definecolor{figsub}{HTML}{586777}
  \definecolor{figgrid}{HTML}{E4E9ED}

  \resizebox{\textwidth}{!}{%
    \begin{tikzpicture}[
      x=1pt,y=1pt,
      font=\fontfamily{lmss}\selectfont,
      title/.style={font=\fontfamily{lmss}\selectfont\bfseries\fontsize{17.5}{21}\selectfont,text=figink,align=center},
      rowlabel/.style={font=\fontfamily{lmss}\selectfont\fontsize{12}{14.5}\selectfont,text=figink,align=right,anchor=east,inner sep=0pt},
      value/.style={font=\fontfamily{lmss}\selectfont\bfseries\fontsize{9.6}{11.6}\selectfont,text=figink,inner sep=0pt},
      legend/.style={font=\fontfamily{lmss}\selectfont\fontsize{12}{14.5}\selectfont,text=figink,anchor=west},
    ]
      \def\labelx{200}
      \def\barx{220}
      \def\barw{330}
      \def\barh{24}
      \def\top{223.2}
      \def\rowgap{47.3}
      \pgfmathsetmacro{\bottomrow}{\top-(3)*\rowgap}%
      \pgfmathsetmacro{\legendy}{\bottomrow-22}%
      \node[title,anchor=center,text width=670pt] at (350,{\top+58}) {Bench-1: Multi-turn text interaction};

      \foreach \pct in {0,25,50,75,100}{%
        \pgfmathsetmacro{\xx}{\barx+\barw*(\pct/100)}%
        \draw[draw=figgrid,line width=.65pt] (\xx,\bottomrow) -- (\xx,\top+5);
      }

      \newcommand{\resultbar}[5]{%
        \pgfmathsetmacro{\yy}{\top-#1*\rowgap}%
        \pgfmathsetmacro{\ww}{\barw*(#3/100)}%
        \pgfmathsetmacro{\tw}{\barw*(#4/100)}%
        \pgfmathsetmacro{\lw}{\barw*(#5/100)}%
        \node[rowlabel,text width=200pt] at (\labelx,\yy+\barh/2) {#2};
        \path[fill=figwin] (\barx,\yy) rectangle ++(\ww,\barh);
        \path[fill=figtie] ({\barx+\ww},\yy) rectangle ++(\tw,\barh);
        \path[fill=figloss] ({\barx+\ww+\tw},\yy) rectangle ++(\lw,\barh);
        \pgfmathsetmacro{\wcx}{\barx+\ww/2}%
        \pgfmathsetmacro{\tcx}{\barx+\ww+\tw/2}%
        \pgfmathsetmacro{\lcx}{\barx+\ww+\tw+\lw/2}%
        \ifdim\ww pt<32pt
          \node[value,anchor=south east] at ({\barx+\ww},{\yy+\barh+3}) {#3\%};
        \else
          \node[value] at (\wcx,{\yy+\barh/2}) {#3\%};
        \fi
        \ifdim\tw pt<32pt
          \node[value,anchor=south east] at ({\barx+\ww+\tw},{\yy+\barh+3}) {#4\%};
        \else
          \node[value] at (\tcx,{\yy+\barh/2}) {#4\%};
        \fi
        \ifdim\lw pt<20pt
          \node[value,anchor=south east] at ({\barx+\ww+\tw+\lw},{\yy+\barh+3}) {#5\%};
        \else
          \ifdim\lw pt<32pt
            \node[value,anchor=east] at
              ({\barx+\ww+\tw+\lw-2},{\yy+\barh/2}) {#5\%};
          \else
            \node[value] at (\lcx,{\yy+\barh/2}) {#5\%};
          \fi
        \fi
      }

      \resultbar{0}{Prompt following\\vs. HappyOyster-Directing}{65.0}{25.0}{10.0}
      \resultbar{1}{Prompt following\\vs. LingBot-World v2}{75.0}{16.7}{8.3}
      \resultbar{2}{Physical plausibility\\vs. HappyOyster-Directing}{45.0}{43.3}{11.7}
      \resultbar{3}{Physical plausibility\\vs. LingBot-World v2}{68.3}{26.7}{5.0}

      \path[fill=figwin] (205,{\legendy-5}) rectangle ++(14,10);
      \node[legend] at (227,\legendy) {\shortname{} win};
      \path[fill=figtie] (385,{\legendy-5}) rectangle ++(14,10);
      \node[legend] at (407,\legendy) {Tie};
      \path[fill=figloss] (470,{\legendy-5}) rectangle ++(14,10);
      \node[legend] at (492,\legendy) {\shortname{} loss};

      \pgfresetboundingbox
      \path[use as bounding box]
        (0,{\legendy-14}) rectangle (700,{\top+78});
    \end{tikzpicture}%
  }
  \caption{
    \textbf{Pairwise human evaluation of multi-turn text control.}
    We compare \shortname{} with HappyOyster-Directing and
    LingBot-World~v2 on prompt following and physical plausibility.
    Within each evaluation dimension, the two 100\% stacked bars show wins
    (green), ties (gray), and losses (pink) from the perspective of
    \shortname{}.
  }
  \label{fig:multi_turn_text_control}
\end{figure*}

\paragraph{Human evaluation.}
For each metric of Hy-Nomas-Bench, we report pairwise win, tie, and loss percentage from the perspective of \shortname{}. The net win rate is the win percentage minus the loss percentage.
Evaluations are conducted by groups of trained human annotators. Each group has three annotators, who discuss and calibrate their scoring criteria to improve consistency. 

\subsection{Comparison with Other World Models}
Overall, \shortname{} demonstrates superior interaction capability and supports different control signals for interaction. Notably, Genie 3 does not provide a free-form text interaction capability in its official API. Happy Oyster adopts separate models for each control signal, thus can not support text and camera interleaved interaction. Although LingBot-World v2 can cover all the settings by adapting its open-source model, it exhibits substantially weaker performance in following free-form text instructions, as its text-based control is implemented by selecting VLM-generated proposals to guide the generation of the subsequent video segment, rather than taking free-form text instructions from the user. From Bench-1 to Bench-3, \shortname{} consistently achieves the best performance, highlighting both its versatility and effectiveness in handling heterogeneous control signals. Despite these advantages, spatial memory remains an area for improvement. While \shortname{} performs comparably to Happy Oyster, it still trails Genie 3 in maintaining long-term spatial consistency, suggesting research opportunities to further improve \shortname{}'s memory capability.
We break down the results for each benchmark below.

\paragraph{Bench-1: Multi-turn free-form text interaction.}
Figure~\ref{fig:multi_turn_text_control} summarizes the results. For prompt following, \shortname{} achieves win rates of 65.0\% against HappyOyster-Directing and 75.0\% against LingBot-World~v2. For physical plausibility, \shortname{} achieves win rates of 45.0\% and 68.3\%.
These results demonstrate that \shortname{} consistently outperforms both models in instruction following and physical plausibility, with stronger advantages over LingBot-World~v2. We hypothesize that our structured captions and attention design contribute to these improvements by providing explicit supervision of actions, scene changes and camera movements. Figures~\ref{fig:qual_text_robot}  provides qualitative examples for multi-turn text interaction.

\begin{figure}[t]
    \centering
    \begingroup
    \newcommand{\QualitativeFont}{\fontfamily{lmss}\selectfont}
    \resizebox{\linewidth}{!}{\input{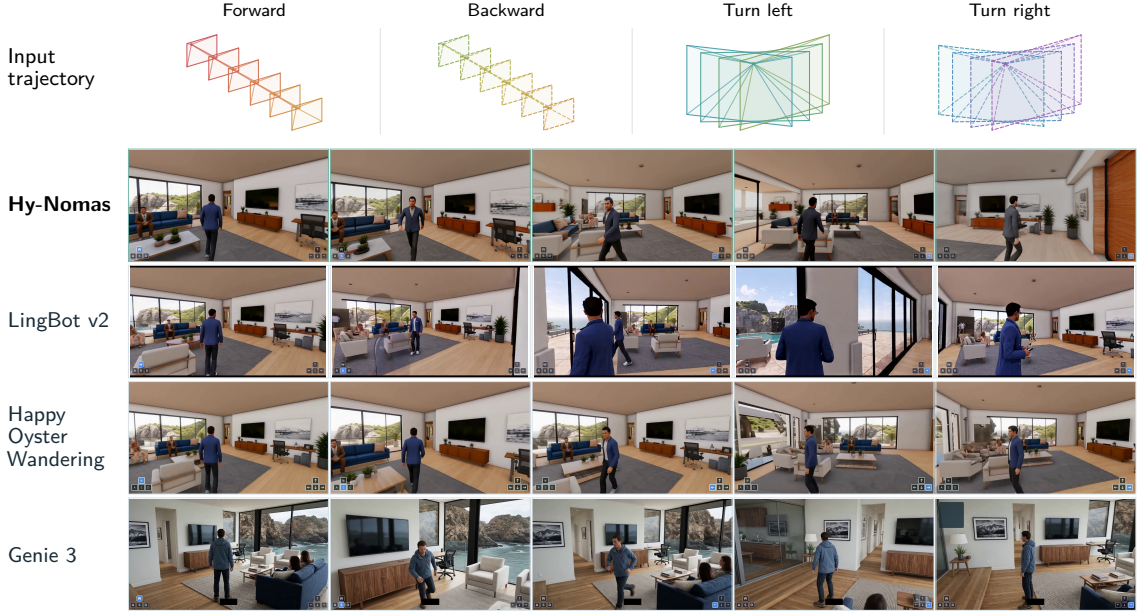}}
    \endgroup
    \caption{\textbf{Qualitative comparison of camera control in an indoor scene.}
Rows show \shortname{} (Ours), LingBot-World~v2, HappyOyster-Wandering,
and Genie~3. 
The comparison illustrates how each model responds to navigation commands
while maintaining a third-person view of the character and the surrounding
room. The middle frame of LingBot-World~v2 shows a duplicated character,
illustrating a scene-consistency failure during navigation.
Visible keyboard indicators denote active controls in the source
recordings.}
    \label{fig:qual_camera}
\end{figure}

\begin{figure*}[t]
  \centering
  \vspace{-5mm}
  \definecolor{figwin}{HTML}{70BC93}
  \definecolor{figtie}{HTML}{E2E6EB}
  \definecolor{figloss}{HTML}{E2A1AF}
  \definecolor{figink}{HTML}{152A3A}
  \definecolor{figsub}{HTML}{586777}
  \definecolor{figgrid}{HTML}{E4E9ED}

  \resizebox{\textwidth}{!}{%
    \begin{tikzpicture}[
      x=1pt,y=1pt,
      font=\fontfamily{lmss}\selectfont,
      title/.style={font=\fontfamily{lmss}\selectfont\bfseries\fontsize{17.5}{21}\selectfont,text=figink,align=center},
      rowlabel/.style={font=\fontfamily{lmss}\selectfont\fontsize{12}{14.5}\selectfont,text=figink,align=right,anchor=east,inner sep=0pt},
      value/.style={font=\fontfamily{lmss}\selectfont\bfseries\fontsize{9.6}{11.6}\selectfont,text=figink,inner sep=0pt},
      legend/.style={font=\fontfamily{lmss}\selectfont\fontsize{12}{14.5}\selectfont,text=figink,anchor=west},
    ]
      \def\labelx{200}
      \def\barx{220}
      \def\barw{330}
      \def\barh{24}
      \def\top{175.89999999999998}
      \def\rowgap{47.3}
      \pgfmathsetmacro{\bottomrow}{\top-(2)*\rowgap}%
      \pgfmathsetmacro{\legendy}{\bottomrow-22}%
      \node[title,anchor=center,text width=670pt] at (350,{\top+58}) {Bench-2: Camera control};

      \foreach \pct in {0,25,50,75,100}{%
        \pgfmathsetmacro{\xx}{\barx+\barw*(\pct/100)}%
        \draw[draw=figgrid,line width=.65pt] (\xx,\bottomrow) -- (\xx,\top+5);
      }

      \newcommand{\resultbar}[5]{%
        \pgfmathsetmacro{\yy}{\top-#1*\rowgap}%
        \pgfmathsetmacro{\ww}{\barw*(#3/100)}%
        \pgfmathsetmacro{\tw}{\barw*(#4/100)}%
        \pgfmathsetmacro{\lw}{\barw*(#5/100)}%
        \node[rowlabel,text width=200pt] at (\labelx,\yy+\barh/2) {#2};
        \path[fill=figwin] (\barx,\yy) rectangle ++(\ww,\barh);
        \path[fill=figtie] ({\barx+\ww},\yy) rectangle ++(\tw,\barh);
        \path[fill=figloss] ({\barx+\ww+\tw},\yy) rectangle ++(\lw,\barh);
        \pgfmathsetmacro{\wcx}{\barx+\ww/2}%
        \pgfmathsetmacro{\tcx}{\barx+\ww+\tw/2}%
        \pgfmathsetmacro{\lcx}{\barx+\ww+\tw+\lw/2}%
        \ifdim\ww pt<32pt
          \node[value,anchor=south east] at ({\barx+\ww},{\yy+\barh+3}) {#3\%};
        \else
          \node[value] at (\wcx,{\yy+\barh/2}) {#3\%};
        \fi
        \ifdim\tw pt<32pt
          \node[value,anchor=south east] at ({\barx+\ww+\tw},{\yy+\barh+3}) {#4\%};
        \else
          \node[value] at (\tcx,{\yy+\barh/2}) {#4\%};
        \fi
        \ifdim\lw pt<32pt
          \node[value,anchor=south east] at ({\barx+\ww+\tw+\lw},{\yy+\barh+3}) {#5\%};
        \else
          \node[value] at (\lcx,{\yy+\barh/2}) {#5\%};
        \fi
      }

      \resultbar{0}{Genie 3}{32.0}{54.0}{14.0}
      \resultbar{1}{HappyOyster-Wandering}{36.0}{54.0}{10.0}
      \resultbar{2}{LingBot-World v2}{29.0}{61.0}{10.0}

      \path[fill=figwin] (205,{\legendy-5}) rectangle ++(14,10);
      \node[legend] at (227,\legendy) {\shortname{} win};
      \path[fill=figtie] (385,{\legendy-5}) rectangle ++(14,10);
      \node[legend] at (407,\legendy) {Tie};
      \path[fill=figloss] (470,{\legendy-5}) rectangle ++(14,10);
      \node[legend] at (492,\legendy) {\shortname{} loss};

      \pgfresetboundingbox
      \path[use as bounding box]
        (0,{\legendy-14}) rectangle (700,{\top+78});
    \end{tikzpicture}%
  }
  \caption{
    \textbf{Pairwise human evaluation of camera control.}
    We compare \shortname{} with Genie~3,
    HappyOyster-Wandering, and LingBot-World~v2 on 100 shared
    trajectories. We measure how accurately the model responds to navigation commands.
  }
  \vspace{-3mm}
  \label{fig:camera_control}
\end{figure*}
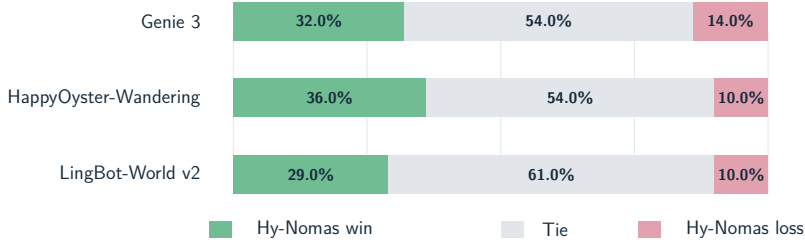

\paragraph{Bench-2: Camera control.}
Figure~\ref{fig:camera_control} presents the results on camera control across all four models. \shortname{} consistently achieves higher win rates than loss rates against the other three models. Compared with the results on Bench-1, the advantage of \shortname{} is smaller for camera control, which has been studied more widely for the world model. Figure~\ref{fig:qual_camera} complements these results with an indoor navigation example. 

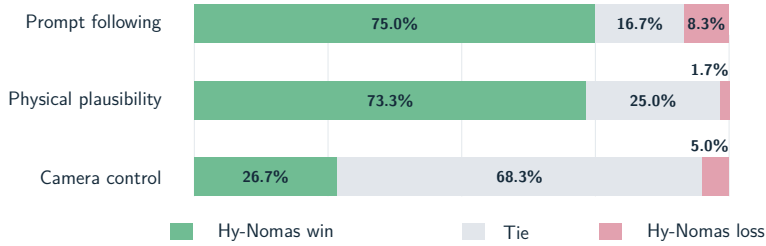
\begin{figure*}[t]
  \centering
  \definecolor{figwin}{HTML}{70BC93}
  \definecolor{figtie}{HTML}{E2E6EB}
  \definecolor{figloss}{HTML}{E2A1AF}
  \definecolor{figink}{HTML}{152A3A}
  \definecolor{figsub}{HTML}{586777}
  \definecolor{figgrid}{HTML}{E4E9ED}

  \resizebox{\textwidth}{!}{%
    \begin{tikzpicture}[
      x=1pt,y=1pt,
      font=\fontfamily{lmss}\selectfont,
      title/.style={font=\fontfamily{lmss}\selectfont\bfseries\fontsize{17.5}{21}\selectfont,text=figink,align=center},
      rowlabel/.style={font=\fontfamily{lmss}\selectfont\fontsize{12}{14.5}\selectfont,text=figink,align=right,anchor=east,inner sep=0pt},
      value/.style={font=\fontfamily{lmss}\selectfont\bfseries\fontsize{9.6}{11.6}\selectfont,text=figink,inner sep=0pt},
      legend/.style={font=\fontfamily{lmss}\selectfont\fontsize{12}{14.5}\selectfont,text=figink,anchor=west},
    ]
      \def\labelx{200}
      \def\barx{220}
      \def\barw{330}
      \def\barh{24}
      \def\top{175.89999999999998}
      \def\rowgap{47.3}
      \pgfmathsetmacro{\bottomrow}{\top-(2)*\rowgap}%
      \pgfmathsetmacro{\legendy}{\bottomrow-22}%
      \node[title,anchor=center,text width=670pt] at (350,{\top+58}) {Bench-3: Interleaved text and camera control};

      \foreach \pct in {0,25,50,75,100}{%
        \pgfmathsetmacro{\xx}{\barx+\barw*(\pct/100)}%
        \draw[draw=figgrid,line width=.65pt] (\xx,\bottomrow) -- (\xx,\top+5);
      }

      \newcommand{\resultbar}[5]{%
        \pgfmathsetmacro{\yy}{\top-#1*\rowgap}%
        \pgfmathsetmacro{\ww}{\barw*(#3/100)}%
        \pgfmathsetmacro{\tw}{\barw*(#4/100)}%
        \pgfmathsetmacro{\lw}{\barw*(#5/100)}%
        \node[rowlabel,text width=200pt] at (\labelx,\yy+\barh/2) {#2};
        \path[fill=figwin] (\barx,\yy) rectangle ++(\ww,\barh);
        \path[fill=figtie] ({\barx+\ww},\yy) rectangle ++(\tw,\barh);
        \path[fill=figloss] ({\barx+\ww+\tw},\yy) rectangle ++(\lw,\barh);
        \pgfmathsetmacro{\wcx}{\barx+\ww/2}%
        \pgfmathsetmacro{\tcx}{\barx+\ww+\tw/2}%
        \pgfmathsetmacro{\lcx}{\barx+\ww+\tw+\lw/2}%
        \ifdim\ww pt<32pt
          \node[value,anchor=south east] at ({\barx+\ww},{\yy+\barh+3}) {#3\%};
        \else
          \node[value] at (\wcx,{\yy+\barh/2}) {#3\%};
        \fi
        \ifdim\tw pt<32pt
          \node[value,anchor=south east] at ({\barx+\ww+\tw},{\yy+\barh+3}) {#4\%};
        \else
          \node[value] at (\tcx,{\yy+\barh/2}) {#4\%};
        \fi
        \ifdim\lw pt<20pt
          \node[value,anchor=south east] at ({\barx+\ww+\tw+\lw},{\yy+\barh+3}) {#5\%};
        \else
          \ifdim\lw pt<32pt
            \node[value,anchor=east] at
              ({\barx+\ww+\tw+\lw-2},{\yy+\barh/2}) {#5\%};
          \else
            \node[value] at (\lcx,{\yy+\barh/2}) {#5\%};
          \fi
        \fi
      }

      \resultbar{0}{Prompt following}{75.0}{16.7}{8.3}
      \resultbar{1}{Physical plausibility}{73.3}{25.0}{1.7}
      \resultbar{2}{Camera control}{26.7}{68.3}{5.0}

      \path[fill=figwin] (205,{\legendy-5}) rectangle ++(14,10);
      \node[legend] at (227,\legendy) {\shortname{} win};
      \path[fill=figtie] (385,{\legendy-5}) rectangle ++(14,10);
      \node[legend] at (407,\legendy) {Tie};
      \path[fill=figloss] (470,{\legendy-5}) rectangle ++(14,10);
      \node[legend] at (492,\legendy) {\shortname{} loss};

      \pgfresetboundingbox
      \path[use as bounding box]
        (0,{\legendy-14}) rectangle (700,{\top+78});
    \end{tikzpicture}%
  }
  \caption{
    \textbf{Pairwise human evaluation of interleaved text and camera control.}
    We compare \shortname{} with the adapted LingBot-World~v2 baseline on
    60 scenarios with alternating text and camera inputs. We consider prompt following, physical plausibility, and camera control during the evaluation.
  }
  \vspace{-5mm}\label{fig:interleaved_control}
\end{figure*}

\paragraph{Bench-3: Interleaved text and camera control.}
\shortname{} natively supports interleaved text and camera control within one unified model, without requiring session resets or switching between specialized models. We compare against LingBot-World~v2 using its open-source model, evaluating its performance under free-form text instructions. As shown in Figure~\ref{fig:interleaved_control}, \shortname{} achieves win rate of 75.0\% and 73.3\% in prompt following and physical plausibility, respectively. For camera control, the two models exhibit more comparable performance, with \shortname{} achieving a win rate of 26.7\%. Overall, these results highlight the effectiveness of \shortname{} in maintaining instruction-following accuracy and physical plausibility under interleaved text and camera control.

\begin{figure*}[t]
  \centering
  \definecolor{figwin}{HTML}{70BC93}
  \definecolor{figtie}{HTML}{E2E6EB}
  \definecolor{figloss}{HTML}{E2A1AF}
  \definecolor{figink}{HTML}{152A3A}
  \definecolor{figsub}{HTML}{586777}
  \definecolor{figgrid}{HTML}{E4E9ED}

  \resizebox{\textwidth}{!}{%
    \begin{tikzpicture}[
      x=1pt,y=1pt,
      font=\fontfamily{lmss}\selectfont,
      title/.style={font=\fontfamily{lmss}\selectfont\bfseries\fontsize{17.5}{21}\selectfont,text=figink,align=center},
      rowlabel/.style={font=\fontfamily{lmss}\selectfont\fontsize{12}{14.5}\selectfont,text=figink,align=right,anchor=east,inner sep=0pt},
      value/.style={font=\fontfamily{lmss}\selectfont\bfseries\fontsize{9.6}{11.6}\selectfont,text=figink,inner sep=0pt},
      legend/.style={font=\fontfamily{lmss}\selectfont\fontsize{12}{14.5}\selectfont,text=figink,anchor=west},
    ]
      \def\labelx{200}
      \def\barx{220}
      \def\barw{330}
      \def\barh{24}
      \def\top{175.89999999999998}
      \def\rowgap{47.3}
      \pgfmathsetmacro{\bottomrow}{\top-(2)*\rowgap}%
      \pgfmathsetmacro{\legendy}{\bottomrow-22}%
      \node[title,anchor=center,text width=670pt] at (350,{\top+58}) {Bench-4: Spatial memory under viewpoint revisitation};

      \foreach \pct in {0,25,50,75,100}{%
        \pgfmathsetmacro{\xx}{\barx+\barw*(\pct/100)}%
        \draw[draw=figgrid,line width=.65pt] (\xx,\bottomrow) -- (\xx,\top+5);
      }

      \newcommand{\resultbar}[5]{%
        \pgfmathsetmacro{\yy}{\top-#1*\rowgap}%
        \pgfmathsetmacro{\ww}{\barw*(#3/100)}%
        \pgfmathsetmacro{\tw}{\barw*(#4/100)}%
        \pgfmathsetmacro{\lw}{\barw*(#5/100)}%
        \node[rowlabel,text width=200pt] at (\labelx,\yy+\barh/2) {#2};
        \path[fill=figwin] (\barx,\yy) rectangle ++(\ww,\barh);
        \path[fill=figtie] ({\barx+\ww},\yy) rectangle ++(\tw,\barh);
        \path[fill=figloss] ({\barx+\ww+\tw},\yy) rectangle ++(\lw,\barh);
        \pgfmathsetmacro{\wcx}{\barx+\ww/2}%
        \pgfmathsetmacro{\tcx}{\barx+\ww+\tw/2}%
        \pgfmathsetmacro{\lcx}{\barx+\ww+\tw+\lw/2}%
        \ifdim\ww pt<32pt
          \node[value,anchor=south east] at ({\barx+\ww},{\yy+\barh+3}) {#3\%};
        \else
          \node[value] at (\wcx,{\yy+\barh/2}) {#3\%};
        \fi
        \ifdim\tw pt<32pt
          \node[value,anchor=south east] at ({\barx+\ww+\tw},{\yy+\barh+3}) {#4\%};
        \else
          \node[value] at (\tcx,{\yy+\barh/2}) {#4\%};
        \fi
        \ifdim\lw pt<32pt
          \node[value,anchor=south east] at ({\barx+\ww+\tw+\lw},{\yy+\barh+3}) {#5\%};
        \else
          \node[value] at (\lcx,{\yy+\barh/2}) {#5\%};
        \fi
      }

      \resultbar{0}{Genie 3}{13.3}{56.7}{30.0}
      \resultbar{1}{HappyOyster}{30.0}{46.7}{23.3}
      \resultbar{2}{LingBot-World v2}{58.3}{40.0}{1.7}

      \path[fill=figwin] (205,{\legendy-5}) rectangle ++(14,10);
      \node[legend] at (227,\legendy) {\shortname{} win};
      \path[fill=figtie] (385,{\legendy-5}) rectangle ++(14,10);
      \node[legend] at (407,\legendy) {Tie};
      \path[fill=figloss] (470,{\legendy-5}) rectangle ++(14,10);
      \node[legend] at (492,\legendy) {\shortname{} loss};

      \pgfresetboundingbox
      \path[use as bounding box]
        (0,{\legendy-14}) rectangle (700,{\top+78});
    \end{tikzpicture}%
  }
  \caption{
    \textbf{Pairwise human evaluation of spatial memory.}
    We compare \shortname{} with Genie~3, HappyOyster, and
    LingBot-World~v2 on 60 shared scenarios involving reversible or
    revisiting camera trajectories. We measure how well the model remembers the anchor object and its surrounding environments.
  }
  \label{fig:spatial_memory}
\end{figure*}
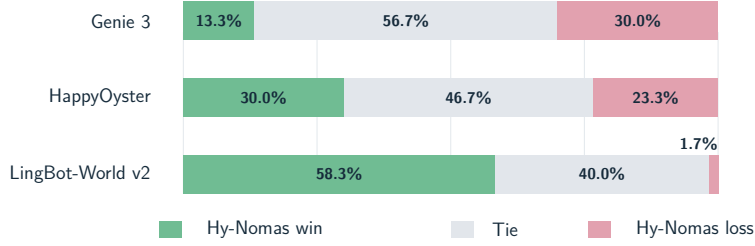

\paragraph{Bench-4: Spatial memory.}
For spatial memory evaluation, we assess whether models can maintain consistent object and scene appearance when revisiting previously observed locations. Human annotators compare the revisited views with their original appearances, focusing on the preservation of scene layout, object identities (\eg, the anchor object), and spatial relationships. A model is considered more spatially consistent if it accurately reconstructs previously observed regions without introducing inconsistent objects, altering scene geometry, or violating spatial relationships.

\begin{figure}[!htbp]
    \centering
    \begingroup
    \newcommand{\QualitativeFont}{\fontfamily{lmss}\selectfont}
    \resizebox{\linewidth}{!}{\input{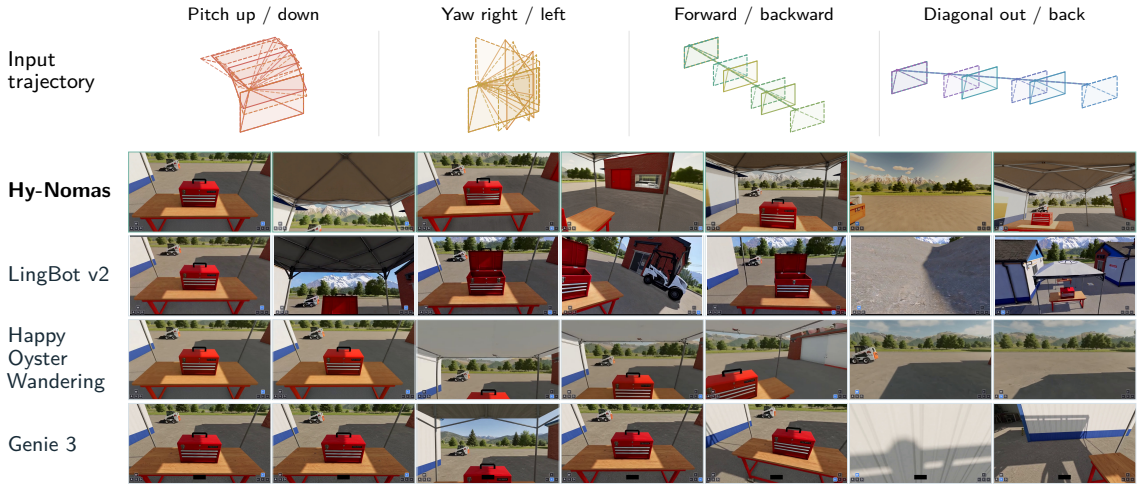}}
    \endgroup
    \caption{\textbf{Qualitative comparison of spatial memory under viewpoint revisitation.}
Rows show \shortname{} (Ours), LingBot-World~v2, HappyOyster-Wandering,
and Genie~3; columns show successive frames from a camera-only rollout.
\shortname{} and Genie~3 retain the closed toolbox in
revisited views, whereas LingBot-World~v2 changes it to an open toolbox.
This case illustrates object-state consistency; aggregate results in
Figure~\ref{fig:spatial_memory} provide the broader comparison.}
    \label{fig:qual_memory}
\end{figure}

Figure~\ref{fig:spatial_memory} presents the results of the spatial memory evaluation. Compared with Genie~3, \shortname{} achieves a win rate of 13.3\%, a tie rate of 56.7\%, and a loss rate of 30.0\%, indicating that Genie~3 maintains stronger spatial consistency. Genie supports grounding in Street View data~\citep{rivas2026genie_streetview}; we hypothesize that may contribute to improved spatial consistency. Against HappyOyster, \shortname{} achieves comparable performance, with 30.0\% wins, 46.7\% ties, and 23.3\% losses. In contrast, \shortname{} substantially outperforms LingBot-World~v2, achieving a win rate of 58.3\% and a loss rate of only 1.7\%. Overall, these results demonstrate the competitive spatial memory capabilities of \shortname{}, while highlighting a remaining gap with Genie~3. We hypothesize that incorporating more data with looped camera trajectories during training could further improve spatial memory consistency. We leave this for future research explorations. Figure~\ref{fig:qual_memory} illustrates a spatial memory example in a scene with a red toolbox on a wooden workbench. Comparing the initial view of the anchor object and surrounding environment, later views make changes to object appearance and scene structure.

\subsection{LoRA Adaption Results}
\label{sec:lora_results}
We merge the LoRA weights trained on the base autoregressive model into the distilled model, which we recommend as the best practice for LoRA adaptation. We refer to the resulting model as the LoRA-adapted model. 

\begin{figure}[ht]
  \centering
  \includegraphics[width=0.9\textwidth]{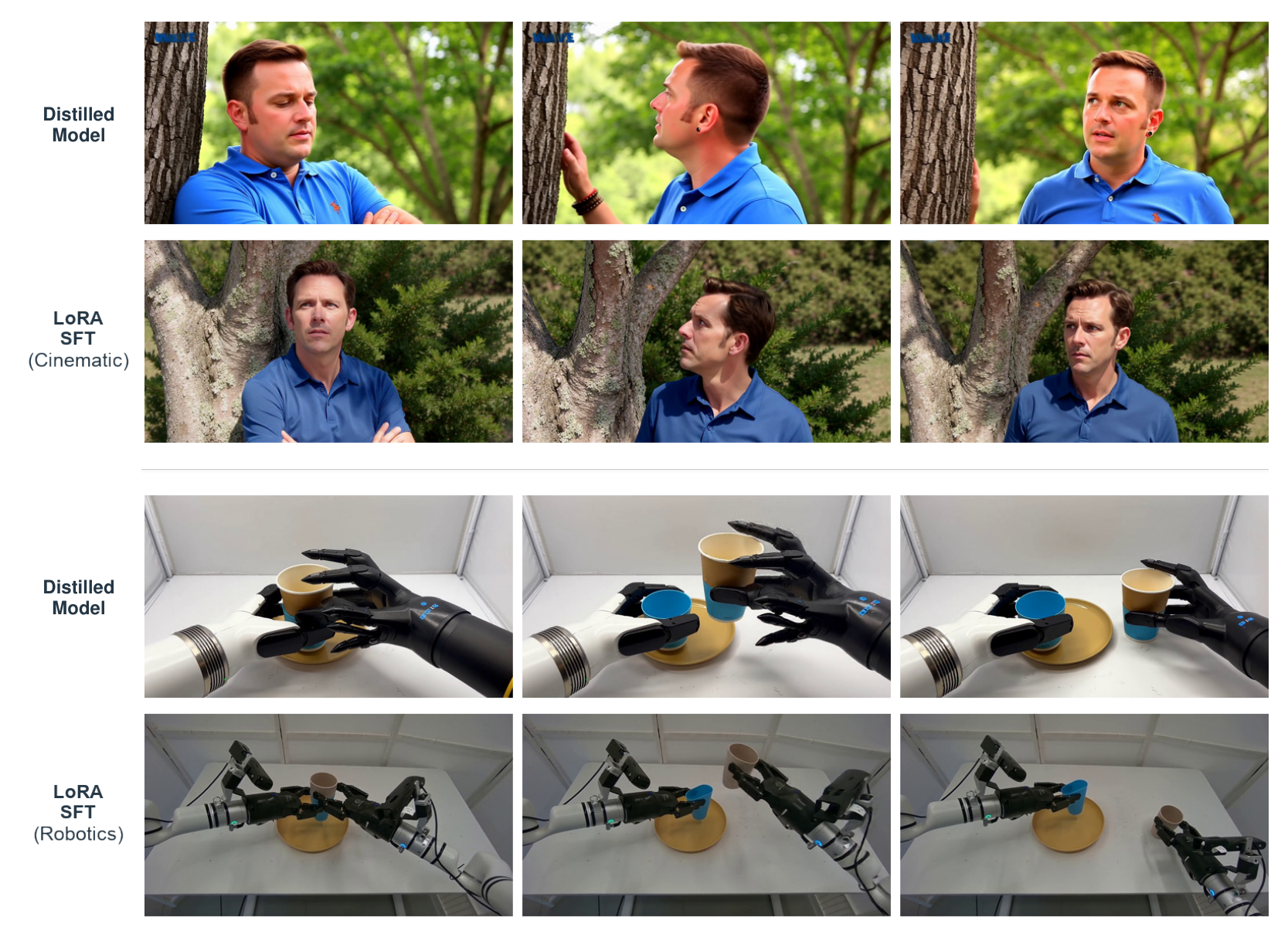}
\vspace{-3mm}  \caption{\textbf{Qualitative results of LoRA-SFT.} We compare the distilled model with domain-specific LoRA-SFT variants. The upper examples show improved cinematic aesthetics after cinematic LoRA-SFT, while the lower examples show more physically plausible robotic manipulation, with cup deformation more faithfully reflecting the gripper's applied pressure. \textbf{LoRA-SFT} denotes the distilled model with the trained LoRA weights merged into it.}
  \label{fig:lora_fig_viz}
\end{figure}

\paragraph{Cinematic LoRA Results.} Compared with the distilled model, the cinematic LoRA-adapted model produces a more visually appealing cinematic style, as illustrated in Figure~\ref{fig:lora_fig_viz}. 
We further evaluate both models on Bench-1 for multi-turn text interaction, assessing prompt following and physical plausibility, and on Bench-2 for camera control, to examine whether the improvements in cinematic visual quality introduce any regressions in other capabilities. 

\begin{figure*}[t]
  \vspace{1mm}
  \centering
  \definecolor{lorawin}{HTML}{70BC93}
  \definecolor{loratie}{HTML}{E2E6EB}
  \definecolor{loraloss}{HTML}{E2A1AF}
  \definecolor{loratext}{HTML}{152A3A}
  \definecolor{lorasub}{HTML}{586777}
  \definecolor{loragrid}{HTML}{E4E9ED}

  \resizebox{\textwidth}{!}{%
    \begin{tikzpicture}[x=1pt,y=1pt,font=\fontfamily{lmss}\selectfont,
      category/.style={font=\fontfamily{lmss}\selectfont\fontsize{15}{18}\selectfont,
        text=loratext,anchor=east,align=right,text width=112pt},
      value/.style={font=\fontfamily{lmss}\selectfont\bfseries\fontsize{12}{14.5}\selectfont,
        text=loratext,inner sep=0pt},
      model/.style={font=\fontfamily{lmss}\selectfont\bfseries\fontsize{15}{18}\selectfont,
        text=loratext},
      paneltitle/.style={font=\fontfamily{lmss}\selectfont\bfseries\fontsize{17.5}{21}\selectfont,
        text=loratext,align=center},
      panelsubtitle/.style={font=\fontfamily{lmss}\selectfont\bfseries\fontsize{15}{18}\selectfont,
        text=loratext,align=center},
      legend/.style={font=\fontfamily{lmss}\selectfont\fontsize{15}{18}\selectfont,
        text=lorasub,anchor=west}]
      \def\panelw{510}
      \def\panelgap{10.56}
      \def\labelx{126}
      \def\barx{134}
      \def\barw{350}
      \def\barh{34}

      \newcommand{\lorabar}[5]{%
        \pgfmathsetmacro{\ww}{\barw*(#3/100)}%
        \pgfmathsetmacro{\tw}{\barw*(#4/100)}%
        \pgfmathsetmacro{\lw}{\barw*(#5/100)}%
        \node[category] at (\labelx,#1) {#2};
        \path[fill=lorawin] (\barx,{#1-\barh/2}) rectangle ++(\ww,\barh);
        \path[fill=loratie] ({\barx+\ww},{#1-\barh/2}) rectangle ++(\tw,\barh);
        \path[fill=loraloss] ({\barx+\ww+\tw},{#1-\barh/2}) rectangle ++(\lw,\barh);
        \pgfmathsetmacro{\wcx}{\barx+\ww/2}%
        \pgfmathsetmacro{\tcx}{\barx+\ww+\tw/2}%
        \pgfmathsetmacro{\lcx}{\barx+\ww+\tw+\lw/2}%
        \ifdim\ww pt<32pt
          \node[value,anchor=south east] at ({\barx+\ww},{#1+\barh/2+3}) {#3\%};
        \else
          \node[value] at (\wcx,#1) {#3\%};
        \fi
        \ifdim\tw pt<32pt
          \node[value,anchor=south east] at ({\barx+\ww+\tw},{#1+\barh/2+3}) {#4\%};
        \else
          \node[value] at (\tcx,#1) {#4\%};
        \fi
        \ifdim\lw pt<32pt
          \node[value,anchor=south east] at ({\barx+\ww+\tw+\lw},{#1+\barh/2+3}) {#5\%};
        \else
          \node[value] at (\lcx,#1) {#5\%};
        \fi
      }

      \newcommand{\loragridlines}[2]{%
        \foreach \pct in {0,50,100}{%
          \pgfmathsetmacro{\xx}{\barx+\barw*(\pct/100)}%
          \draw[draw=loragrid,line width=.65pt] (\xx,#1) -- (\xx,#2);
        }%
      }

\begin{scope}[xshift=0pt]
  \node[paneltitle,text width=490pt] at (255,300)
    {Bench-1: Multi-turn text interaction};
  \node[panelsubtitle,text width=490pt] at (255,272)
    {vs. Distilled model};
  \loragridlines{112}{238}
  \lorabar{210}{Prompt following}{35.0}{41.0}{24.0}
  \lorabar{140}{Physical plausibility}{46.0}{41.0}{13.0}
\end{scope}

\begin{scope}[xshift={\panelw+\panelgap}]
  \node[paneltitle,text width=490pt] at (255,300)
    {Bench-2: Camera control};
  \node[panelsubtitle,text width=490pt] at (255,272)
    {vs. Distilled model};
  \loragridlines{158}{192}
  \lorabar{175}{Camera control}{19.0}{44.0}{37.0}
\end{scope}

      \path[fill=lorawin] (252,64) rectangle ++(14,10);
      \node[legend] at (274,69) {LoRA-SFT win};
      \path[fill=loratie] (447,64) rectangle ++(14,10);
      \node[legend] at (469,69) {Tie};
      \path[fill=loraloss] (572,64) rectangle ++(14,10);
      \node[legend] at (594,69) {LoRA-SFT loss};

      \pgfresetboundingbox
      \path[use as bounding box] (0,48) rectangle (1030.56,330);
    \end{tikzpicture}%
  }
  \vspace{-6mm}
  \caption{
    \textbf{Pairwise human evaluation of cinematic LoRA-SFT against the distilled model.}
    We evaluate prompt following and physical plausibility on Bench-1,
    and camera control on Bench-2.
    The three 100\% stacked bars show wins (green), ties (gray),
    and losses (pink) from the perspective of LoRA-SFT
    relative to the base model. \textbf{LoRA-SFT} denotes the distilled model with the trained LoRA weights merged into it.
  }

    \label{fig:lora_control_memory}
  \vspace{-2mm}
\end{figure*}
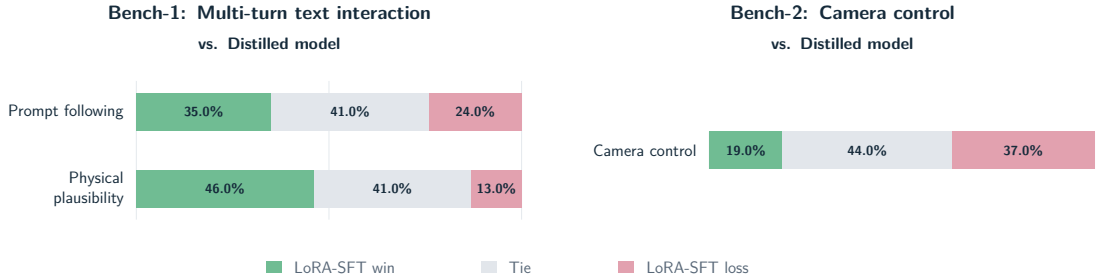

Quantitative results are reported in Figure~\ref{fig:lora_control_memory}. Human evaluation demonstrates a strong preference for the adapted model in terms of fewer visual artifacts and improved text-prompt following. On Bench-1, the LoRA-SFT model shows improvements in both prompt following and physical plausibility, achieving win rates of $35\%$ and $46\%$, compared to loss rates of $24\%$ and $13\%$, respectively. However, we observe a moderate regression in camera-control performance, with the LoRA-SFT model achieving a $19\%$ win rate versus a $37\%$ loss rate. We hypothesize that this degradation stems from the limited diversity of camera trajectories in the SFT dataset, which may weaken previously learned camera-control capabilities, particularly for underrepresented camera motions. These results highlight the effectiveness of LoRA in adapting visual styles with minimal training data, while underscoring the importance of preserving diverse control signals during fine-tuning. 

\paragraph{Robotics LoRA Results.} Compared to the distilled model as a baseline, illustrated in Figure~\ref{fig:lora_fig_viz}, the merged robotics-specific distilled checkpoint can perform out-of-distribution dual-arm tasks, while the baseline confuses the generation of dual-arm and human hands due to the heavy training set distribution of human egocentric data.

\subsection{PE Results}
\begin{figure*}[t]
  \vspace{-7mm}
  \centering
  \definecolor{pesbg}{HTML}{FFFFFF}
  \definecolor{peswin}{HTML}{70BC93}
  \definecolor{pestie}{HTML}{E2E6EB}
  \definecolor{pesloss}{HTML}{E2A1AF}
  \definecolor{pestext}{HTML}{152A3A}
  \definecolor{pessub}{HTML}{586777}
  \definecolor{pesgrid}{HTML}{E4E9ED}
  \resizebox{\textwidth}{!}{%
    \begin{tikzpicture}[x=1pt,y=1pt,font=\fontfamily{lmss}\selectfont,
      pescategory/.style={font=\fontfamily{lmss}\selectfont\fontsize{15}{18}\selectfont,
        text=pestext,anchor=east,align=right,text width=112pt},
      pesvalue/.style={font=\fontfamily{lmss}\selectfont\bfseries\fontsize{12}{14.5}\selectfont,
        text=pestext,inner sep=0pt},
      pestick/.style={font=\fontfamily{lmss}\selectfont\fontsize{15}{18}\selectfont,text=pessub},
      pesmodel/.style={font=\fontfamily{lmss}\selectfont\bfseries\fontsize{15}{18}\selectfont,text=pestext}]
      \def\barw{181}
      \newcommand{\pessegment}[5]{%
        \path[fill=#4] (#1,{#2-17}) rectangle ({#1+#3},{#2+17});
        \ifdim #3pt<35pt
          \ifdim #1pt<135pt
            \node[pesvalue,anchor=south west] at
              (#1,{#2+20}) {#5\%};
          \else
            \ifdim #1pt>319pt
              \ifdim #1pt<321pt
                \node[pesvalue,anchor=south west] at
                  (#1,{#2+20}) {#5\%};
              \else
                \node[pesvalue,anchor=south east] at
                  ({#1+#3},{#2+20}) {#5\%};
              \fi
            \else
              \node[pesvalue,anchor=south east] at
                ({#1+#3},{#2+20}) {#5\%};
            \fi
          \fi
        \else
          \node[pesvalue] at ({#1+#3/2},#2) {#5\%};
        \fi
      }
      \newcommand{\pesstack}[8]{%
        \pgfmathsetmacro{\pestotal}{#3+#4+#5}%
        \pgfmathsetmacro{\pesww}{\barw*(#3/\pestotal)}%
        \pgfmathsetmacro{\pestw}{\barw*(#4/\pestotal)}%
        \pgfmathsetmacro{\peslw}{\barw-\pesww-\pestw}%
        \pgfmathsetmacro{\peststart}{#1+\pesww}%
        \pgfmathsetmacro{\peslstart}{#1+\pesww+\pestw}%
        \pessegment{#1}{#2}{\pesww}{peswin}{#6}
        \pessegment{\peststart}{#2}{\pestw}{pestie}{#7}
        \pessegment{\peslstart}{#2}{\peslw}{pesloss}{#8}
      }
      \newcommand{\pesgridpanel}[5]{%
        \node[pesmodel] at ({#1+\barw/2},#5) {vs. #2};
        \foreach \pct in {0,50,100}{%
          \pgfmathsetmacro{\pesxx}{#1+\barw*(\pct/100)}%
          \draw[draw=pesgrid,line width=0.65pt] (\pesxx,#3) -- (\pesxx,#4);
        }
      }
      \begin{scope}[xshift=0pt]
        \path[fill=pesbg,rounded corners=9pt] (0,12) rectangle (510,342);
        \node[font=\fontfamily{lmss}\selectfont\bfseries\fontsize{17.5}{21}\selectfont,text=pestext]
          at (255,306) {Bench-1: Multi-turn text interaction};
        \pesgridpanel{134}{Qwen3.8-27B}{110.5}{209.5}{237.5}
        \pesgridpanel{320}{Kimi-K3}{110.5}{209.5}{237.5}
        \node[pescategory] at (126,192.5) {Prompt following};
        \node[pescategory] at (126,127.5) {Physical plausibility};
        \pesstack{134}{192.5}{24}{21}{13}
          {41.4}{36.2}{22.4}
        \pesstack{134}{127.5}{19}{24}{11}
          {35.2}{44.4}{20.4}
        \pesstack{320}{192.5}{22}{18}{17}
          {38.6}{31.6}{29.8}
        \pesstack{320}{127.5}{12}{30}{12}
          {22.2}{55.6}{22.2}
      \end{scope}
      \begin{scope}[xshift=520.56pt]
        \path[fill=pesbg,rounded corners=9pt] (0,12) rectangle (510,342);
        \node[font=\fontfamily{lmss}\selectfont\bfseries\fontsize{17.5}{21}\selectfont,text=pestext]
          at (255,306) {Bench-3: Interleaved text and camera control};
        \pesgridpanel{134}{Qwen3.8-27B}{78}{242}{270}
        \pesgridpanel{320}{Kimi-K3}{78}{242}{270}
        \node[pescategory] at (126,225) {Prompt following};
        \node[pescategory] at (126,160) {Physical plausibility};
        \node[pescategory] at (126,95) {Camera control};
        \pesstack{134}{225}{23}{24}{17}
          {35.9}{37.5}{26.6}
        \pesstack{134}{160}{14}{30}{15}
          {23.7}{50.8}{25.4}
        \pesstack{134}{95}{7}{45}{8}
          {11.7}{75.0}{13.3}
        \pesstack{320}{225}{17}{24}{21}
          {27.4}{38.7}{33.9}
        \pesstack{320}{160}{14}{24}{22}
          {23.3}{40.0}{36.7}
        \pesstack{320}{95}{4}{43}{12}
          {6.8}{72.9}{20.3}
      \end{scope}
      \path[fill=peswin] (254.28,43) rectangle ++(16,16);
      \node[pestick,anchor=west] at (279.28,51) {Hy-Nomas-PE win};
      \path[fill=pestie] (449.28,43) rectangle ++(16,16);
      \node[pestick,anchor=west] at (474.28,51) {Tie};
      \path[fill=pesloss] (574.28,43) rectangle ++(16,16);
      \node[pestick,anchor=west] at (599.28,51) {Hy-Nomas-PE loss};
      \pgfresetboundingbox
      \path[use as bounding box] (0,30) rectangle (1030.56,350);
    \end{tikzpicture}%
  }
  \caption{
    \textbf{Rank-gap-weighted human evaluation of Hy-Nomas-PE (ours).}
    Bench-1 and bench-3 are reported separately;
    each pair compares Hy-Nomas-PE with Qwen3.8-27B and Kimi-K3.
    Bench-1 evaluates prompt following and physical plausibility; bench-3 also
    evaluates camera control. Category labels are shared within each benchmark.
    Each 100\% stacked bar shows win (green), tie (gray), and loss (pink)
    votes from the perspective of Hy-Nomas-PE.
  }
    \vspace{-5mm}
    \label{fig:pe_sft_mp_joint_win_tie_loss}
\end{figure*}
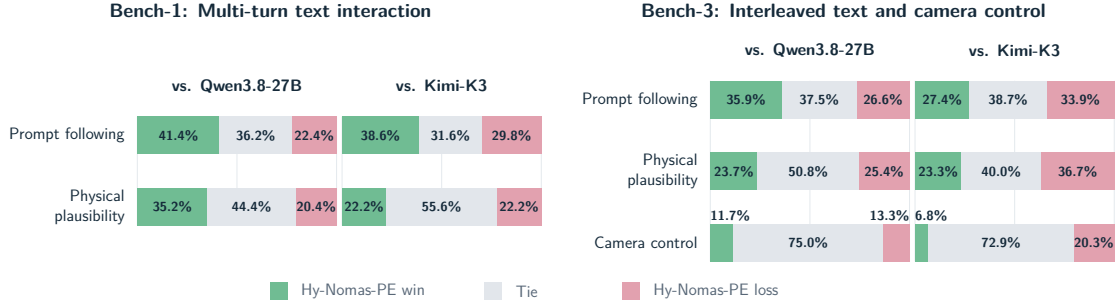
We also evaluate the performance of our PE model by end-to-end directly comparison. We consider three different models: Hy-Nomas-PE, the original Qwen3.8-27B, and Kimi-K3. Figure~\ref{fig:pe_sft_mp_joint_win_tie_loss} presents the evaluation results on Bench-1 and Bench-3. Compared to Qwen3.8-27B, our PE achieves substantial improvements in both prompt following and physical plausibility while maintaining the camera control performance, significantly narrowing the performance gap with Kimi-K3. Besides, our PE achieves a nearly 100\% success rate in adhering to the required response format for user inputs, compared to approximately 50\% for Qwen3.8-27B and 70\% for Kimi-K3.

\section{Related Work}
\paragraph{World models}
The definition of world model now encompasses a broad range of objectives and approaches. Representation-based approaches, such as V-JEPA~\citep{bardes2024revisiting}, learn predictive visual features as a foundation for video understanding. Geometry-based approaches, including systems from World Labs~\citep{worldlabs2025worlds}, emphasize persistent 3D representations that support rendering and navigation. Reconstruction models such as VGGT~\citep{wang2025vggt} provide complementary building blocks for recovering scene geometry from visual observations. Video-based approaches model world evolution through generated observations, drawing on large collections of videos to learn visual dynamics and action-conditioned transitions~\citep{bruce2024genie,yang2023unisim}. A central question is whether learning from visual appearances and motion alone is sufficient to capture the underlying physical dynamics, or whether additional structural priors and supervision are required. Controlled studies have revealed limitations in the physical generalization of video generation models, even as model capacity and training data scale~\citep{kang2025physical}. Meanwhile, recent work has demonstrated that post-training techniques, such as DPO~\citep{wallace2024diffusion,rafailov2023direct}, can improve physical consistency~\citep{cai2026phygdpo}. Nevertheless, whether further scaling of data and model capacity, combined with reinforcement learning (RL), can fundamentally overcome these limitations and enable video-generation-based world models to learn robust physical dynamics remains an open research question. \\

In this sense, world modeling brings together several long-standing goals of computer vision: 3D reconstruction, video understanding, and visual generation~\citep{ren2025gen3c,ren2025videoworld,bahmani2024ac3d}. It also raises a broader question: can intelligence emerge from perceiving the visual world? A complementary direction uses a language- or vision-language-model agent to write and refine Blender code for constructing or editing 3D scenes, as explored in BlenderAlchemy~\citep{huang2024blenderalchemy}. Such a pipeline could additionally incorporate generative video editing to enhance visual realism. This design places planning and program synthesis in an agent, while delegating scene execution and rendering to graphics tools. It illustrates a possible division of responsibilities, without establishing where intelligence must reside. Whether multimodal learning contributes distinct forms of intelligence beyond providing interfaces for perception and generation remains an open question. \\

Across these approaches, world models offer promising applications in autonomous driving~\citep{hu2023gaia,wang2023drivedreamer,ren2025cosmosdrivedreams,du2023unipi,nvidia2025cosmos}, robotics and action-conditioned simulation~\citep{yang2023unisim,zhou2024robodreamer,gao2026dreamdojo,nvidia2026cosmos3,jang2025dreamgen}, and controllable game generation~\citep{bruce2024genie,feng2026matrix,li2025hunyuan,valevski2024gamengen,che2024gamegenx,tang2025gamecraft2}. \shortname{} pursues the video-based direction, motivated by the scalability and diversity of available video data. Our focus is on turning this visual experience into interactive generation, where user actions shape subsequent observations and the evolving state of the world.

\paragraph{Video diffusion models}
Diffusion models have become a major approach to video generation~\citep{ho2022video,blattmann2023align,kodaira2026streamdit,hacohen2025ltx,wan2025open,chen2024videocrafter2,blattmann2023stable} that extend image diffusion models~\citep{rombach2022high,saharia2022photorealistic,ramesh2022hierarchical,dai2023emu,betker2023improving,podell2023sdxl,openai2025nativeimage,bfl2025kontext,esser2024scaling,cao2025hunyuanimage} to temporally coherent video synthesis through spatiotemporal modeling and joint image--video training. Diffusion Transformers~\citep{peebles2023scalable} establish a scalable backbone for latent diffusion, while Sora~\citep{openai2024sora} and MovieGen~\citep{polyak2024movie} illustrate the potential of large-scale video generation using spacetime latent patches. Seedance~\citep{seedance2026seedance2} further advances visual quality and the commercial potential of video generation. In comparison, HunyuanVideo~1.5~\citep{wu2025hunyuanvideo}, the backbone of our model, provides an efficient 8B-parameter foundation that balances visual quality with computational accessibility. Beyond fixed-length generation, Diffusion Forcing~\citep{chen2024diffusionforcing} combines causal sequence prediction with independently sampled noise levels, supporting flexible generation horizons. Self Forcing~\citep{huang2025selfforcing} addresses the training--inference mismatch in autoregressive video diffusion by training on model-generated histories with KV caching. Building on these directions, \shortname adapts a pretrained video generator to interactive, segment-wise generation. Its temporally aligned captions, and segment-wise and action-aware attention connect user instructions to specific portions of the rollout, while causal KV caching retains context across successive interactions.

\paragraph{Video-based world models}
Video-based world models extend video generation toward action-conditioned, interactive environments~\citep{hong2025relic,sun2025worldplay}. Research in this area emphasizes several complementary dimensions, including physical plausibility for physical AI, responsive controllability, and persistent world state. Beyond generating visually convincing sequences, an interactive world model must respond to actions introduced during a rollout and preserve their consequences in subsequent observations. \\

Genie~\citep{bruce2024genie} learns controllable environments from videos without ground-truth action annotations through a latent action model and autoregressive dynamics. Genie~2~\citep{deepmind2024genie2} and Genie~3~\citep{deepmind2025genie3} further develop action-controllable environments, with Genie~3 emphasizing real-time interaction and improved visual consistency. LingBot-World~v2~\citep{gao2026infinite} explores long-horizon causal generation, more diverse action-control events, and agentic control through pilot
and director components. HappyOyster~\citep{happyoyster2026} provides interactive world creation through Directing and Wandering modes, which we evaluate for text interaction and camera navigation, respectively. \\

These approaches share challenges in responsive control, physical plausibility, spatial memory, and efficient long-horizon generation. Within this direction, \shortname focuses on native alternation between free-form textual instructions and continuous camera control in one 8B model. It combines temporally structured supervision, persistent causal context, and a VLM-based director that translates user intent into segment-level prompts and generation durations. Our evaluation examines text interaction, camera control, interleaved control, and spatial memory separately to characterize these complementary capabilities.

\section{Discussion and Future Directions}
\label{sec:discussion}

\paragraph{Gaming and robotics.}
Gaming and robotics place different demands on world models. For gaming, a practical challenge is how quickly the model responds to a player's input. Turning the camera or issuing an action should produce a visual response with minimal delay, while repeated interactions should preserve scene consistency. Distillation can reduce the number of denoising steps~\citep{yin2024dmd2,zhao2026causalforcingpp,liu2026rolling}, and quantization can reduce memory use and computational cost~\citep{chen2026longlive,feng2025qvdit}. These gains must be evaluated alongside visual quality and consistency over long rollouts~\citep{xu2026worldmark,xu2026worldroambench}. For robotics, the challenge extends beyond generating a plausible response: the predicted outcome must match the effect of an executable robot action~\citep{zhu2024irasim,wang2026interactive}. For example, a grasp prediction should reflect whether the gripper makes contact with an object and successfully lifts it. World-action models such as DreamZero~\citep{ye2026worldaction} jointly predict actions and future visual states, providing a potential route toward this capability. Adapting \shortname{} to robotic control would require training on videos paired with robot actions and evaluating whether its predictions support successful task execution. Datasets such as Open X-Embodiment~\citep{khazatsky2024droid,openx2024embodiment} provide a starting point, although coverage across tasks, environments, and robot embodiments remains limited. As an initial step in this direction, we demonstrate LoRA-based fine tuning on robotics adaptation.

\paragraph{Agentic world models.}
Long-horizon world generation requires remembering previously observed scenes to maintain consistency when users revisit them. Retaining the full KV cache on the GPU becomes increasingly costly as rollouts grow. One possible approach is to offload older context to CPU memory or persistent storage and selectively retrieve information relevant to the next segment~\citep{yu2025contextmemory,yu2026memlearner}. When a user returns to a room, for example, the system could retrieve earlier views of that room rather than attend to the entire interaction history. WorldMem~\citep{xiao2025worldmem} demonstrates the value of retrieving historical observations for consistency when revisiting scenes.

Different types of information may benefit from different memory representations: geometry can encode spatial relationships, reference images can preserve visual appearance, and text can summarize events and state changes, such as a door being opened or an object being moved~\cite{huang2025memory}. In our spatial-memory experiments, explicitly describing the expected scene content when revisiting a location can improve consistency. An agent could use this mechanism to specify which objects should reappear, where they should be, and what has changed since the previous visit, while visual references could help preserve their precise appearance. Cached keys and values offer another form of memory by retaining features that the generator can directly reuse, although managing their GPU memory footprint requires offloading, compression, or eviction. A practical agentic system would coordinate these memory representations, retrieve relevant information, and update stored records as the world evolves, ensuring that outdated observations do not override the current scene state. \\

Retrieval alone, however, cannot account for changes that occur outside the field of view. Consider a wall clock: after the user leaves a room and returns ten minutes later, the clock should remain in the same place with the same appearance, but display a time ten minutes later. Reusing an earlier image would recover its appearance without updating its time. An agent could instead record the clock's identity, location, and observed time, compute its updated state from elapsed simulation time, and pass that state together with a reference image to the generator. Inspired by memory management in language-model agents~\citep{packer2023memgpt}, this approach would expose memory retrieval, state updates, prompt rewriting, and video generation as tools. LingBot-World~v2~\citep{gao2026infinite} provides an encouraging example of agent coordination for interactive world generation. An important next step is to test whether such coordination can maintain explicit world states and render them faithfully over long interactions.

\paragraph{Data and captioning.}
Despite having only 8B parameters, \shortname{} achieves stronger text-driven action control than the larger models evaluated in our benchmark. We hypothesize that accurate, detailed captions contribute to this result by explicitly describing changes over time, which reduces the burden on the diffusion transformer to "imagine" unspecified content and allows it to devote more capacity to just rendering the described scenes and actions correctly. This hypothesis is consistent with findings that improved captions strengthen prompt following in image generation~\citep{betker2023improving}. These observations suggest a potential direction for future systems: using highly detailed prompts to specify scene content and temporal evolution, with diffusion models serving primarily as renderers of these descriptions. Such an approach places greater demands on the quality and specificity of both training captions and inference-time prompts, making caption and Prompt Enhancer central challenges. Data curation should therefore consider not only visual quality but also whether the caption accurately describes the video. For example, a clip containing motion blur during a rapid camera turn may still provide useful supervision if the caption describes both the camera motion and the resulting blur. Similarly, explicitly captioning a cut or an editing effect may reduce ambiguity about the observed transition. 

\paragraph{Infrastructure and scaling.}
Higher resolution and longer temporal context increase the number of video tokens, raising training costs, inference latency, and KV-cache memory requirements. Efficient attention, sequence parallelism, and cache management are therefore critical to extending rollouts and enabling interactive generation. These infrastructure advances also make it more practical to explore larger world models. Recent progress in video generation, exemplified by Seedance~2.0~\citep{seedance2026seedance2}, motivates further investigation into the relationship between model capacity and generation quality. MAGI-2 Preview~\citep{sandai2026magi2preview} explores this direction through a mixture-of-experts architecture (MoE), scaling total capacity while limiting the number of parameters activated per token. A central question is whether such investments translate into better world-model capabilities, including actions following, predicting physically consistent outcomes, and preserving scene details when revisiting locations after long rollouts. These capabilities require direct evaluation and cannot be inferred from improvements in visual quality alone.

\section{Contributors}
All contributors of \shortname are listed in alphabetical order by their last names.

\textbf{Core Contributors:}  Qiaoling Chen, Jiaxiang Cheng, Xueqing Deng, Tiankai Hang, Ji Hou, Jiaqi Li, Yifan Liu, Tao Lu, Junshu Tang, Yu Tang, Yao Teng, Yu Tian, Heng Wang, Weizhi Wang, Xiyao Wang, Xilun Wu, Mingyu Yang, Yangbin Yu, Rui Zhao

\textbf{Contributors:} Longjun Cao, Linghao Chen, Haoyi Duan, Jiaming Han, Mingji Han, Enhua Jiang, Shunkai Li, Yanzhe Liang, Qinwei Ma, Cong Wang, Lei Wang, Bing Wu, Ke Xing, Ray Ying, Zhentao Yu, Key Zhang, Yuan Zhou, Zhengguang Zhou, Kai Zou

\textbf{Acknowledgments:} Xiang Bai, Yuhang Deng, Ying Dong, Qiying Gong, Weiting Guo,  Changrong Hu, Yuxin Jiang, Kysunwang, Zhuohao Li, Sen Liang, Chenguo Lin, Juntian Liu, Yezhou Liu, Yuhao Liu, Yuhao Lu, Chunyi Ma, Kai Song, Yonghao Tan, Renjia Wei, Yifan Wu, Yinhe Wu, Yan Zeng, Weiqiao Zheng

\textbf{Leadership Acknowledgments:} Xudong Lin, Yonglong Tian

\bibliography{colm2024_conference}
\bibliographystyle{colm2024_conference}

\end{document}